\documentclass[pdflatex,sn-mathphys-num,iicol]{sn-jnl}

\usepackage{graphicx}
\graphicspath{{./}}
\usepackage{multirow}
\usepackage{amsmath,amssymb,amsfonts}
\usepackage{booktabs}
\usepackage{textcomp}
\usepackage{url}
\usepackage{xurl}
\makeatletter
\g@addto@macro\UrlBreaks{\do\-}
\makeatother
\usepackage{array}
\usepackage{tabularx}
\AtBeginDocument{\hypersetup{hidelinks}}

\begin{document}

\title[Stop to Decide]{Stop to Decide: Latency-Aware Proprioceptive Navigation Primitives for Mapping-Free Quadruped Inspection}

\author[1]{\fnm{Hanting} \sur{Suo}}
\equalcont{These authors contributed equally to this work.}

\author[2]{\fnm{Haonan} \sur{Yan}}
\equalcont{These authors contributed equally to this work.}

\author[3]{\fnm{Liang} \sur{Wang}}

\author*[1]{\fnm{Aiguo} \sur{Song}}\email{a.g.song@seu.edu.cn}

\affil*[1]{\orgdiv{School of Instrument Science and Engineering}, \orgname{Southeast University}, \orgaddress{\city{Nanjing}, \country{China}}}

\affil[2]{\orgdiv{School of Electrical Engineering}, \orgname{Southeast University}, \orgaddress{\city{Nanjing}, \country{China}}}

\affil[3]{\orgdiv{School of Mechanical Engineering}, \orgname{Southeast University}, \orgaddress{\city{Nanjing}, \country{China}}}

\abstract{Onboard quadruped inspection systems often share limited compute between perception and navigation, reducing the rate at which event-triggered controllers evaluate proprioceptive signals. We study this latency in stair-summit detection and propose a climb--settle ``stop-to-decide'' cadence for structured, mapping-free inspection. On a Unitree Go2, the integrated stair loop ran at $\approx$15~Hz. On a three-level stepped platform whose 50~cm top was shorter than the robot, continuous-climb overshoot increased with per-period advance $v/f$, whereas the climb--settle cadence held observed overshoot near zero (22/45 vs 1/45 pooled over $\approx$30/20/15~Hz; Fisher $p\approx2.4\times10^{-7}$). A logistic dose--response model gives a model-based critical rate of $\approx$19~Hz at 0.30~m/s; a pre-specified 40~Hz held-out check was consistent with the protocol-clean fit. We integrated the detector with line following and a three-segment 90$^\circ$ corridor maneuver in a fully onboard, learning-free stack using an IMU, foot-force sensing, three 1-D ranges, and one line camera. The corridor maneuver completed 20/20 trials without contact, compared with 14/20 completions and 12 wall contacts for in-place yaw; the full course completed 18/20 trials. Results are limited to one calibrated course, robot, and operator but identify loop rate as a deployment parameter for proprioceptive event detection.}

\keywords{quadruped robots, autonomous inspection, mapping-free navigation, stair traversal, control-loop latency, proprioceptive sensing}

\maketitle

\section{Introduction}

Industrial inspection is a major deployment context for legged robots. Quadrupeds such as ANYmal and Boston Dynamics Spot have been demonstrated at offshore platforms, oil-and-gas terminals, geothermal plants, electrical converter stations, and fusion-research facilities, where recurring tasks include traversing stairs, grating walkways, narrow service corridors, and painted or taped guide lines between equipment cabinets \cite{gehring2021field,bellicoso2018advances,anybotics_oilgas,bd_nationalgrid,bd_jpower,bd_austria,ori2024jet}. Surveys of robotic inspection in buildings, construction, and infrastructure identify sensor cost, payload, and perception-stack computational complexity as important adoption barriers \cite{halder2023construction,ha2023robots}. These constraints are especially relevant when the deployment target is a commercial quadruped on which the engineering team must build a reproducible, onboard, and low-maintenance autonomy stack.

A natural question is whether the perceptive sensing layer typically assumed by these autonomy stacks is strictly necessary for the structured-inspection segment of the deployment envelope. Recent quadruped-navigation systems aimed at the same hardware regime --- for example the ROS 2 + SLAM-Toolbox + Nav2 pipelines reported on Spot and on the Unitree Go2 EDU platform \cite{becoy2025autonomous} --- rely on full 3D LiDAR point clouds, RGB-D depth, or fused multi-modal perception to localize and to detect terrain transitions. In parallel, a substantial body of \emph{locomotion-level} work has shown that proprioceptive signals alone --- joint state, body-mounted IMU, and contact sensing --- already carry enough information for stable walking across mud, rubble, snow, and stairs of arbitrary rise/run, including in the explicit "blind" formulation of model-based controllers \cite{bledt2018cheetah3,dicarlo2018dynamic} and of learning-based policies \cite{lee2020learning,kumar2021rma,miki2022learning,chamorro2024blind}. Yet these two literatures rarely meet: proprioception-only work is overwhelmingly about \emph{how to keep walking}, not about \emph{where to go, when to switch behavior, and how to chain sub-tasks into a route}. The navigation layer is typically delegated upstream to an exteroceptive stack.

This paper investigates the design corner that results when both ends of the literature are pulled together: can the navigation layer itself --- route execution, behavior switching, multi-segment task chaining --- be implemented under a strict mapping-free and learning-free sensing budget on a commercial quadruped, for the structured-inspection case? The platform is a Unitree Go2 EDU with an NVIDIA Jetson Orin onboard computer, running C++ controllers through Unitree's \texttt{unitree\_sdk2} DDS interface rather than a ROS 2 runtime. The sensing budget includes only the robot's built-in IMU (pitch, roll, yaw, angular velocity), four \texttt{foot\_force} ground-contact sensors, and the front/left/right 1D distance topics published on \texttt{rt/utlidar/range\_info}. A monocular camera is used for line-following, but only for extracting the painted track centerline; it is not used for obstacle detection, depth estimation, terrain segmentation, or scene understanding. The stack runs end-to-end onboard with no off-board computation, no LiDAR point cloud, no RGB-D depth, and no learned policy.

To test whether such a stack can complete an inspection-style course, we built a physical track that combines three recurring structured-environment features: (i) a painted line that the robot follows through several heading changes, (ii) a 55 cm narrow corridor in which the robot's 75 cm $\times$ 35 cm standing body footprint and swing-leg envelope make in-place yaw rotation unreliable, and (iii) a multi-level stepped platform whose top is shorter than the robot, so that a late summit-arrival decision causes an overshoot and a fall. Corridor traversal at width margins approaching the footprint diagonal has been addressed through continuous whole-body optimization with control-barrier functions \cite{liao2023walking} or morphological compression of the body \cite{buchanan2019walking,buchanan2021perceptive}. By contrast, we did not find a controlled study that isolates onboard summit-arrival overshoot as a function of navigation-loop rate in the proprioception-only stair literature. Our system combines deterministic sub-controllers under one upper-level state machine, targeting a lightweight and deployable operating point rather than peak performance or generality.

The evaluation has three parts. First, the stair ablation crosses cadence with imposed loop rate and shows that continuous-climb overshoot increases with per-period advance $x=v/f$, whereas the climb--settle cadence keeps overshoot near zero across the tested rates (Section 6.4). Second, the corridor experiment compares the three-segment maneuver with in-place yaw and measures completion, wall contact, traversal time, and exit-heading error (Section 6.3). Third, end-to-end trials test whether the resulting stack can execute the complete calibrated route (Section 6.1). The detailed counts, confidence intervals, and pre-specified tests are reported with the corresponding experiments rather than used to imply generality beyond the evaluated platform and geometry.

The contributions of this paper are:

\begin{enumerate}
\item \textbf{A characterization of control-loop latency as a failure mechanism for in-motion proprioceptive summit detection, and a climb--settle cadence that suppresses the observed overshoot.} In a controlled ablation crossing cadence with imposed loop rate (arrival rule and all other constants fixed, $n=15$ per cell), overshoot under continuous climbing rises with the per-period advance $v/f$ while the cadence holds it near zero across the tested rates (pooled Fisher $p\approx2.4 \times 10^{-7}$, rate-stratified CMH $p\approx1.9 \times 10^{-6}$; Section 6.4). A logistic dose--response model in $x=v/f$ summarizes the failure and provides a model-based operating rule for the tested geometry (Section 4.5). The ablation isolates the cadence; it does not separately ablate the arrival rule against a single absolute threshold (Section 7.4).

\item \textbf{A geometric feasibility analysis of 90$^\circ$ corner turns in width-constrained corridors, and the three-segment maneuver primitive it justifies.} Closed-form bounds show that contact-free in-place yaw of the nominal swept envelope is blocked even exploiting the corner junction whenever the corridor width is below $\approx$0.85$\cdot$$\sqrt{L^{2}+W^{2}}$, while the deployed 45$^\circ$--push--45$^\circ$ template reduces the required width by $\approx$20\% (Section 4.3); hardware trials corroborate the analysis (20/20 contact-free versus 14/20 with 12 wall contacts; Section 6.3).

\item \textbf{A fully onboard, mapping-free and learning-free navigation stack that integrates these primitives with classical line-following under a single upper-level state machine, evaluated end-to-end} on a physical inspection course (18/20 trials, 77.30 $\pm$ 4.40 s), with the sensing budget reduced to the platform's built-in IMU, four foot-force channels, three 1-D ranges, and one single-task line camera (Sections 3, 6.1).
\end{enumerate}

The constituent locomotion and vision primitives are classical, and we claim no component-level algorithmic novelty: PD line-following is mature \cite{bhuiyan2014vision,engin2012path,krishnamurthy2023vision}, the K-turn template is the canonical narrow-passage maneuver for non-holonomic vehicles \cite{reeds1990optimal}, and IMU-based stair cues date to the RHex line \cite{johnson2011autonomous,wenger2015semi,ilhan2020autonomous}. The contribution lies in the latency failure characterization and cadence remedy (Contribution 1), the corner feasibility analysis (Contribution 2), and an end-to-end demonstration of the resulting deterministic navigation layer under the stated sensing budget (Contribution 3). Section 2 positions the work against the proprioception-only locomotion, stair-traversal, confined-space, and inspection literatures. Section 3 presents the onboard architecture and upper-level state machine. Section 4 describes the sub-controllers, the corner feasibility analysis (4.3), and the latency dose--response model (4.5). Section 5 documents the course and trial protocol; Section 6 reports results; Section 7 discusses limitations and deployment implications.

\section{Related Work}

\subsection{Proprioceptive locomotion and stair-state detection}

Proprioception-only locomotion is mature. Model-based controllers climb debris-littered stairs blind on Cheetah 3 \cite{bledt2018cheetah3,dicarlo2018dynamic}, and the torque-controlled ANYmal \cite{hutter2016anymal} established the platform lineage; learning-based policies extend blind locomotion to mud, snow, rubble, and stairs of arbitrary rise \cite{lee2020learning,kumar2021rma,chamorro2024blind,chen2024slr}, optionally restoring exteroception with graceful degradation to proprioception \cite{miki2022learning}; Ha et al. \cite{ha2025learning} survey this landscape. Throughout, however, the proprioception-only constraint is applied at the \emph{locomotion} layer --- gait, balance, terrain following --- while the navigation layer (route execution, behavior switching, task chaining) is assumed to be solved upstream by vision or LiDAR.

The closest prior to our stair detector is the RHex line. Open-loop stair climbing \cite{moore2002reliable} and proprioceptive compliance adaptation \cite{eich2009adaptive} established exteroception-free traversal; Johnson et al. \cite{johnson2011autonomous} introduced a stationary ``pitch wiggle'' as a stair-finding signal, Wenger et al. \cite{wenger2015semi} used the IMU to detect arrival at a stair landing, and Ilhan, Johnson, and Koditschek \cite{ilhan2020autonomous} composed reactive controllers with perceptually triggered transitions across a corpus of ten stairwells. These studies establish inertial event detection for stair traversal, but do not report a controlled cadence-by-loop-rate evaluation of summit-arrival overshoot. The present study therefore focuses on the narrower question of how an arrival condition behaves when it is evaluated during motion by a rate-limited navigation loop, compared with evaluation after a stop-to-decide settling phase. Recent stair work on the same Go-family hardware uses RGB-based learned detection \cite{wozniak2025rgb}, an orthogonal modality, while gait-level policies \cite{margolis2023walk} and trajectory-synthesis methods \cite{zamani2018stable,bratta2022optimization,barasuol2024stair} assume the stair state is given and address how to keep walking.

Foot-force and tactile sensing have been used for per-leg contact estimation \cite{han2023contact}, touchdown adaptation \cite{zhao2022design}, and --- closest to our corroboration channel --- the vision-free Aliengo strategy of Wang et al. \cite{wang2023climbing}, which classifies ground versus stairs from swing-foot touchdown timing and contact force. None of these works reads the \emph{simultaneous all-four-feet} force topology as evidence of occupancy of a small terminal platform, the role it plays (outside the control path) in Section 4.6. Multi-level stepped structures whose terminal platform is shorter than the robot receive little attention in the surveyed literature, and we found no prior treatment of summit-arrival overshoot under compute-constrained control-loop latency.

\subsection{Legged maneuvering in confined spaces}

Confined-space legged work attacks width-constrained corridors with continuous machinery. The closest comparator is Liao et al. \cite{liao2023walking}, in which a 0.32 m wide quadruped walks through a 0.5 m corridor using whole-body NMPC with exponential discrete control-barrier functions, fed by two depth cameras and a tracking camera registered into an Octomap occupancy grid. Buchanan et al. \cite{buchanan2019walking,buchanan2021perceptive} handle confined traversal --- ~70 cm gaps, 60 cm overhangs --- by perceptive whole-body planning with morphological compression, and reinforcement-learning controllers address narrow tunnels \cite{xu2024dexterous} and pipe inspection \cite{guo2024narrow} with learned policies and visual or privileged inputs. The kinematic alternative we deploy is classical: Reeds and Shepp \cite{reeds1990optimal} enumerate the multi-segment switchback families for non-holonomic vehicles, and the turn--push--turn template of Section 4.3 is the canonical K-turn, which we do not claim as a kinematic contribution. Zhu et al. \cite{zhu2021terrain} characterize in-place quadruped spinning, but not the regime in which the corridor envelope precludes spinning altogether. Section 4.3 gives that regime a closed-form treatment --- when in-place turning is geometrically blocked, including at the corner junction itself, and how much width the discrete maneuver recovers --- which we did not find stated in the legged literature.

\subsection{Quadruped inspection systems and positioning}

Industrial inspection is the most-cited commercial driver for quadrupeds. ANYmal performed autonomous inspection of an offshore HVDC platform \cite{gehring2021field} following the ARGOS challenge line \cite{bellicoso2018advances}, and commercial deployments at oil-and-gas, power, and grid sites are documented in ANYbotics and Boston Dynamics case studies \cite{anybotics_oilgas,bd_nationalgrid,bd_jpower,bd_austria}. Surveys identify sensor cost, payload, and perception-stack complexity as the dominant adoption barriers \cite{halder2023construction,ha2023robots}. The recurring route geometry in these deployments --- painted guide lines, narrow service corridors, stairs --- is the geometry of our course. The line-guided segment reuses mature line-following and lane-detection techniques, including frame-to-frame slope memory for lost-line recovery \cite{bhuiyan2014vision,krishnamurthy2023vision,borkar2012lane,narote2018review}; we claim no contribution there (Section 4.1) beyond noting that gait-induced body roll perturbs the ROI geometry these wheeled-platform priors assume fixed.

This work occupies a specific corner of the design space spanned by four reference classes: continuous-optimization confined-space methods \cite{liao2023walking}, morphological compression \cite{buchanan2019walking,buchanan2021perceptive}, perceptive navigation stacks on identical commercial hardware (ROS 2 + Nav2 + SLAM Toolbox on the Go2 EDU \cite{becoy2025autonomous}), and proprioception-only locomotion \cite{lee2020learning,kumar2021rma,miki2022learning,chamorro2024blind}. Each trades something different for its generality --- solver complexity, morphological degrees of freedom, or a LiDAR-and-map sensing layer --- and none reports the cadence-by-loop-rate question treated in Sections 4.4--4.5. Table 1 locates the present system against representative systems from these classes; Contribution 3 of Section 1 is the end-to-end evaluation of this lightweight operating point on a calibrated structured-inspection route.

\begin{table*}[t]\centering
\caption{\textbf{Positioning of this work against representative legged-navigation systems.} "Integrated" denotes line-following, narrow-corridor traversal, and multi-level stair traversal coordinated under a single navigation stack; "onboard only" denotes no off-board computation. "---" denotes not applicable or not the focus of that work; "n/r" denotes not reported.}
\footnotesize\setlength{\tabcolsep}{3pt}\begin{tabularx}{\textwidth}{@{}>{\raggedright\arraybackslash}p{2.1cm} >{\raggedright\arraybackslash}p{2.6cm} >{\raggedright\arraybackslash}p{1.7cm} >{\raggedright\arraybackslash}X >{\raggedright\arraybackslash}p{0.9cm} >{\raggedright\arraybackslash}p{2.1cm} >{\raggedright\arraybackslash}p{0.9cm}@{}}
\toprule
System (platform) & Navigation sensing & Map / learned policy & Demonstrated geometry (as reported) & \shortstack{On-\\board} & Integrated task chain & \shortstack{Robot\\trial} \\
\midrule
Liao et al. \cite{liao2023walking} (quadruped) & 2$\times$ depth + tracking camera & Octomap / No & 0.5 m corridor, 0.32 m-wide robot & Yes & No (corridor only) & Yes \\
Becoy et al. \cite{becoy2025autonomous} (Unitree Go2 EDU) & 3D LiDAR & SLAM occupancy / No & open indoor environment & Yes & No (open-environment nav) & Yes \\
Buchanan et al. \cite{buchanan2019walking,buchanan2021perceptive} (multilegged) & depth + whole-body planning & Yes / No & ~0.7 m gaps, 0.6 m overhangs & Yes & No (confined traversal) & Yes \\
Proprioceptive RL locomotion \cite{lee2020learning,kumar2021rma,miki2022learning,chamorro2024blind} & IMU + joint state (depth in \cite{miki2022learning}) & No / Yes & natural terrain, stairs (rise n/r) & Yes & No (locomotion only) & Yes \\
RGB / gait stair work \cite{wozniak2025rgb,margolis2023walk} (Go-family) & RGB camera \cite{wozniak2025rgb} / proprioception \cite{margolis2023walk} & No / Yes (\cite{margolis2023walk}) & stairs (n/r) & Yes & No (stairs only) & Yes \\
Wang et al. \cite{wang2023climbing} (Aliengo) & IMU + foot-force & No / No & stairs (lift height n/r) & Yes & No (stairs only) & Yes \\
\textbf{This work (Unitree Go2 EDU)} & IMU + 4$\times$ foot-force + 3$\times$ scalar range + mono camera (line) & \textbf{No / No} & \textbf{0.55 m corridor; 3-level pyramid, 50$\times$50 cm top, 15 cm risers} & \textbf{Yes} & \textbf{Yes} & \textbf{Yes} \\
\bottomrule
\end{tabularx}
\end{table*}

\section{System Architecture}

This section presents the integrated onboard architecture before the individual sub-controllers are described in Section 4. The intent is to make explicit, in one place, the hardware platform, the sensing budget, the runtime stack, and the upper-level state machine that orchestrates behavior switching across the inspection course. The reader may treat this section as the system block that the next section unpacks.

\subsection{Hardware and compute}

The robot is a Unitree Go2 EDU quadruped. The measured standing body dimensions used in the course-design analysis are approximately 75 cm in length and 35 cm in width; the footprint diagonal and swing-leg envelope are therefore larger than the 55 cm corridor width used in the narrow-corridor experiment. The onboard computer is an NVIDIA Jetson Orin module, mounted on the robot and powered from the platform battery. There is no off-board computation: the autonomy stack runs end-to-end on the Jetson, communicating with the Go2 base controller through the manufacturer's \texttt{unitree\_sdk2} C++ interface over Unitree's DDS communication layer. All experiments reported in this paper are run with the robot tethered only by a passive safety leash for fall protection; no wired or wireless tether carries control commands or sensor data.

\subsection{Software stack}

The autonomy stack is implemented as a set of C++ controllers integrated through \texttt{unitree\_sdk2}, Unitree DDS channels, and OpenCV for the line-extraction camera stream. The integrated competition program is a single coordinator that activates the line-following, obstacle/corridor, stair, platform, and terminal-action routines in sequence; the stair detector evaluated in the ablation (Section 6.4) is the climb--settle pitch-hysteresis detector implemented in that program. The implementation choices favor onboard deployability: no ROS 2 runtime is required in the experimental environment, no SLAM library is included, no global map is maintained, and no learned-policy inference engine is loaded.

\subsection{Sensing budget}

The system uses three classes of onboard signal, listed below in declining order of central role in the navigation logic.

\begin{enumerate}
\item \textbf{Inertial measurement unit (IMU).} The Go2's built-in IMU exposes body pitch, roll, yaw, and the corresponding angular velocities. Pitch and the stability of the pitch signal drive the multi-level stair detector of Section 4.4. Yaw is integrated to close the discrete yaw segments of the corridor maneuver of Section 4.3.

\item \textbf{Foot-force contact sensors.} Each of the four legs reports a contact force value at the foot; the four values are read as a length-4 vector \texttt{foot\_force[4]}. The full topology of this vector --- and in particular the simultaneous condition "all four entries above threshold" --- is logged throughout each trial; it forms one component of the overshoot criterion (Section 5.4) and is computed from a channel that the stair detector's own control decision (Section 4.4), which uses only body pitch, does not consume.

\item \textbf{One-dimensional range topics.} The robot publishes the topics \texttt{rt/utlidar/range\_info} carrying front, left, and right one-dimensional distance readings. These are used by the obstacle-detection logic of Section 4.2 and the corridor-detection logic of Section 4.3. We state for precision that these three scalar ranges are exposed by the Go2's onboard L1 LiDAR driver through the \texttt{rt/utlidar/range\_info} topic: the robot does carry a LiDAR at the hardware level, and we read three scalar distances from its driver. We deliberately consume \emph{only} these three scalar ranges and never the LiDAR point cloud, occupancy grid, or any SLAM output. The system does \emph{not} subscribe to LiDAR point clouds and does not maintain an occupancy grid or any other map representation. We therefore describe the stack as \emph{mapping-free} --- it constructs and uses no metric map --- rather than as literally LiDAR-free at the hardware level.
\end{enumerate}

In addition, a monocular camera mounted on the robot's front carries the visual stream consumed by the line-following module of Section 4.1. The camera is used as a single-task line-extraction sensor --- a small region of interest in the lower half of the image is processed for the painted track centerline --- and is not used for obstacle detection, depth estimation, terrain segmentation, semantic understanding, or any cross-task perception. The obstacle-avoidance and stair-traversal sub-controllers are independent of the camera.

Sensors and modalities \textbf{explicitly excluded} from the present system include: 3D LiDAR point clouds, RGB-D depth (Intel RealSense, Orbbec, or equivalent), stereo depth, semantic segmentation networks, learned end-to-end policy controllers, external motion capture, GPS, and any off-board computation. We use the descriptor \emph{mapping-free and learning-free navigation stack} throughout this paper to reflect this exclusion set precisely; the monocular line camera is consistent with the descriptor and is not an exception to it.

\subsection{Upper-level state machine}

The behavior of the integrated stack is governed by a single upper-level state machine running on the Jetson. Each state authorizes exactly one sub-controller to drive the robot, and transitions between states are triggered by signals that are either internal to the active sub-controller (a \emph{self-completion} transition, e.g., the line-follower handing off when an obstacle is detected ahead) or by a global supervisory check (e.g., the obstacle-detection module observing the corridor geometry of Section 4.3 in the range topics and forcing a switch to the narrow-corridor maneuver). The seven states are:

\begin{itemize}
\item \texttt{LINE\_FOLLOW} --- active line-following with PD steering and slope-extrapolation recovery (Section 4.1).
\item \texttt{OBSTACLE\_DETECTED} --- transient state in which the upper-level coordinator decides between open-space arc avoidance and the narrow-corridor maneuver based on the lateral range topics.
\item \texttt{AVOIDANCE} --- open-space arc avoidance (Section 4.2).
\item \texttt{NARROW\_CORRIDOR} --- three-segment maneuver (Section 4.3).
\item \texttt{LINE\_REACQUIRE} --- transient state attempting to recover the painted line after avoidance or corridor traversal.
\item \texttt{STAIR\_TRAVERSAL} --- the internal multi-phase stair finite-state machine of Section 4.4 (\texttt{UP  ->  SUMMIT\_STOP  ->  [SUMMIT\_BACK  ->  SUMMIT\_STOP2]  ->  SUMMIT\_SHIFT  ->  SUMMIT\_TURN  ->  DOWN  ->  EXIT}, the bracketed re-centering phases optional).
\item \texttt{EXIT} --- terminal state on completion of the full course.
\end{itemize}

Figure 1 (course overview), Figure 2 (system block diagram with data flow), and Figure 3 (upper-level state machine transition diagram) accompany this section. The internal stair machine is shown separately in Section 4.4 to avoid confusing the upper-level transitions with the within-stair transitions.
\begin{figure}[t]\centering
\includegraphics[width=\linewidth]{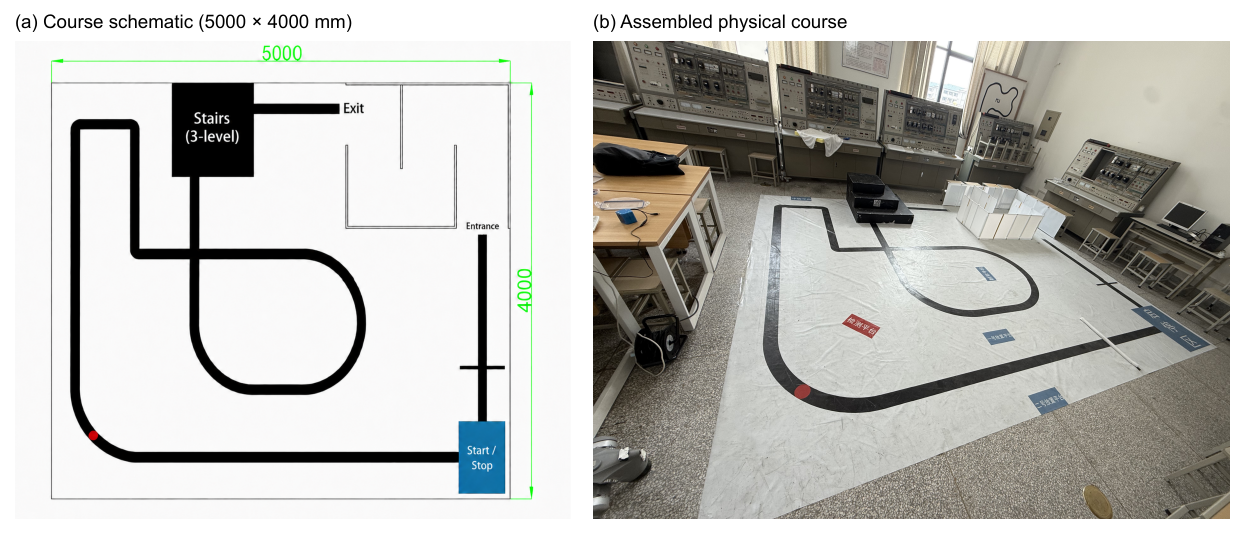}
\caption{\textbf{Assembled inspection course.} (a) Top-down schematic (5000 $\times$ 4000 mm) of the taped route and the three studied segments --- line-following, narrow-corridor avoidance, and three-level stair traversal; (b) photograph of the assembled physical course.}
\end{figure}
\begin{figure*}[t]\centering
\includegraphics[width=0.86\textwidth]{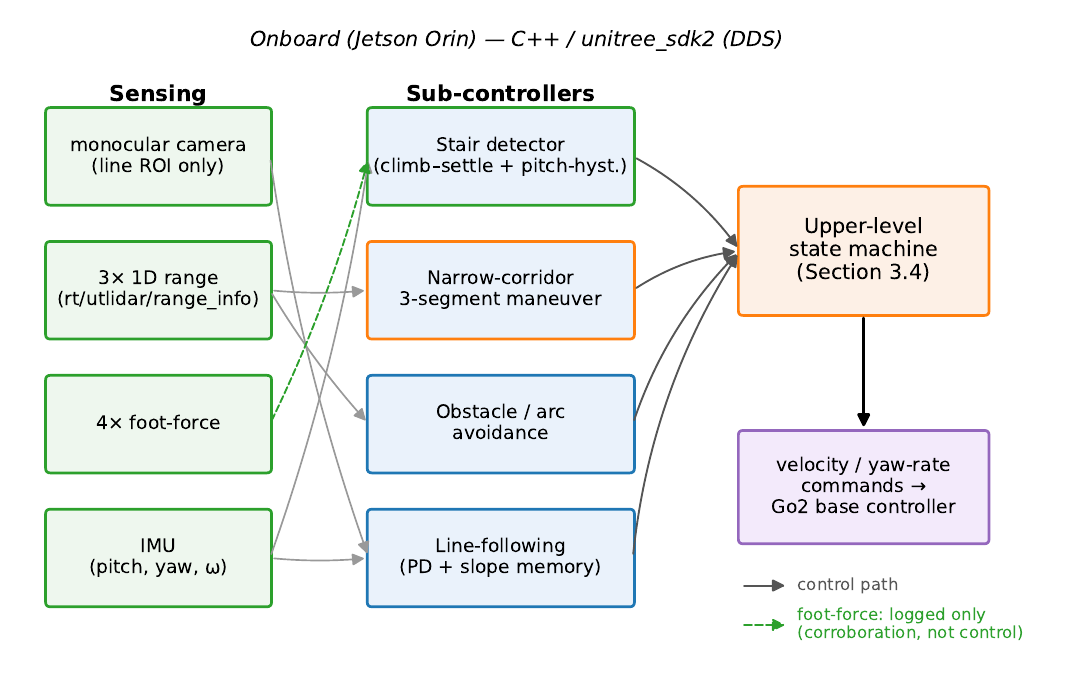}
\caption{\textbf{System block diagram.} Onboard sensing feeds deterministic sub-controllers coordinated by a single upper-level state machine; the four foot-force channels are logged but lie off the control path.}
\end{figure*}
\begin{figure*}[t]\centering
\includegraphics[width=0.9\textwidth]{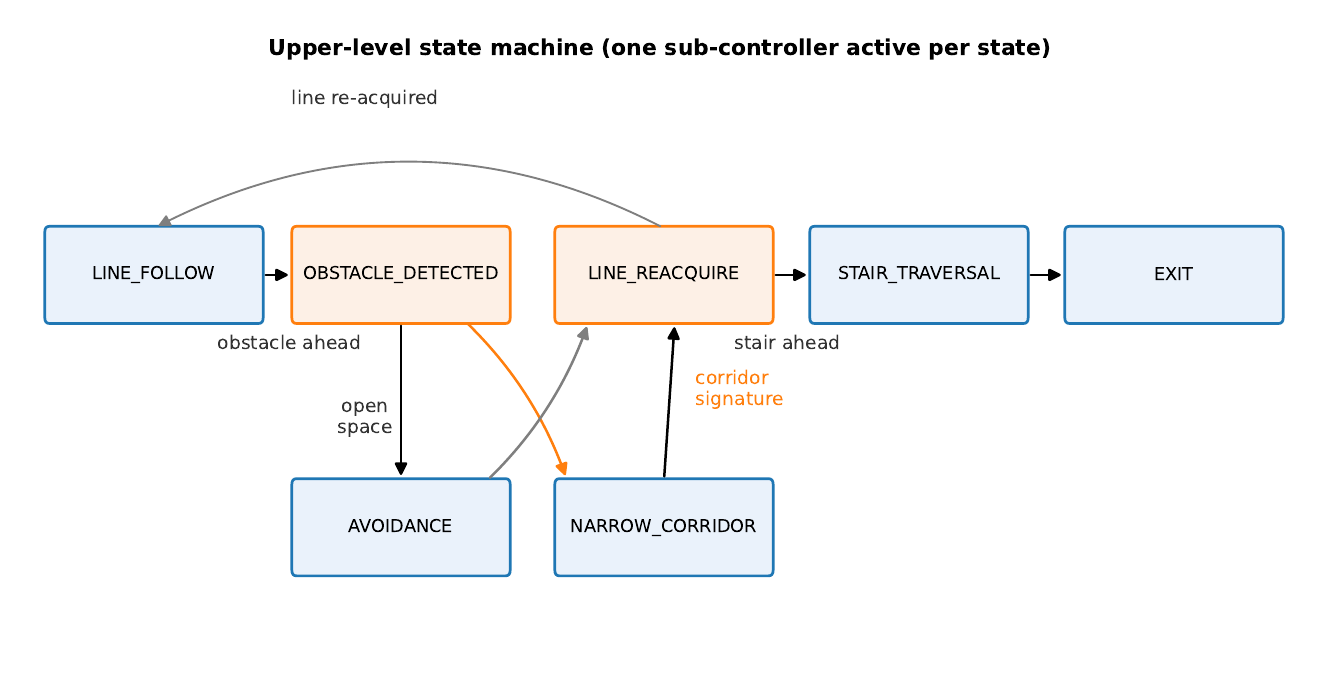}
\caption{\textbf{Upper-level state machine.} Each state authorizes exactly one sub-controller; transitions are triggered by sub-controller self-completion or supervisory checks on the 1D range topics.}
\end{figure*}

\subsection{Timing and runtime assumptions}

All sub-controllers run at the publication rate of the topics they consume. The IMU and \texttt{foot\_force} topics are published at the rate exposed by \texttt{unitree\_sdk2} on the Go2 EDU (in the order of 100 Hz); the range topics and the visual stream are published at lower rates. No sub-controller in the system requires synchronous fusion of asynchronous topics, and all triggers are evaluated on the most recent sample of each subscribed topic. The system assumes the robot is initialized in a stable standing pose at the start of the course and that the upper-level state machine begins in \texttt{LINE\_FOLLOW}.

\section{Methods: Sub-Controller Design}

This section describes the sub-controllers coordinated by the upper-level state machine of Section 3.4, together with the geometric feasibility analysis behind the corridor maneuver (Section 4.3), the latency dose--response model of the stair detector (Section 4.5), and the foot-force corroboration channel (Section 4.6). Subsections 4.3--4.5 instantiate the contributions of this paper; Subsections 4.1 and 4.2 are documented for completeness, and their design choices are re-uses of established techniques.

\subsection{Line-following module (\texttt{line.cpp v3})}

The line-following module is the classical ROI + PD pipeline \cite{bhuiyan2014vision,krishnamurthy2023vision}: the painted centerline is extracted as a horizontal pixel offset in a lower-half region of interest and mapped by a proportional--derivative law to a yaw-rate command, with forward velocity held constant under the normal trot gait. Lost-line recovery carries the most recent valid slope and intercept forward to extrapolate the line position during dropout --- a standard lane-detection technique \cite{borkar2012lane,narote2018review}, retained without a Kalman state because gait-induced body roll perturbs the ROI geometry the wheeled-platform priors assume fixed. We claim no contribution in this module.

\subsection{Range-based obstacle detection and open-space avoidance (\texttt{avoid.cpp})}

Obstacle handling consumes only the three 1-D distance topics on \texttt{rt/utlidar/range\_info}. When the front range falls below a trigger threshold the coordinator enters \texttt{OBSTACLE\_DETECTED} and compares the lateral ranges against a \emph{corridor signature} --- left and right both below a corridor threshold --- to choose between open-space avoidance and the narrow-corridor maneuver of Section 4.3. Open-space avoidance is a constant-radius arc detour away from the side with less clearance until the front range recovers, followed by \texttt{LINE\_REACQUIRE}. No planning, map, or optimization is involved; the behavior exists only to keep the integrated course completable and carries no contribution claim.

\subsection{Narrow-corridor three-segment maneuver (\texttt{avoid\_split.cpp})}

This subsection instantiates the second contribution of the paper. \textbf{Geometric feasibility of the corner turn.} The constraint can be stated in closed form. A 90$^\circ$ in-place yaw of a rectangular body of length $L$ and width $W$ sweeps a disc of diameter $\sqrt{L^2+W^2}$; with the Go2's measured standing footprint ($L \approx 0.75$ m, $W \approx 0.35$ m) the swept diameter is $\approx 0.83$ m, exceeding the 0.55 m corridor width. The corner junction itself does not rescue the maneuver: for two corridor legs of equal width $c$ meeting at a right angle, the largest disc inscribed in the junction region has diameter $2(2-\sqrt{2})\,c \approx 1.17\,c$, so a full in-place rotation is geometrically blocked \emph{anywhere in the junction} whenever $c < \sqrt{L^2+W^2}/(2(2-\sqrt{2})) \approx 0.85\,\sqrt{L^2+W^2}$ --- for this platform, whenever $c < 0.71$ m. The 0.55 m corridor therefore blocks the in-place turn by a 0.16 m margin. We state the precise scope of this argument: it guarantees that the swept envelope of the nominal footprint violates the free-space boundary, not that every physical attempt fails --- a compliant robot may survive wall contact, consistent with the in-place-yaw baseline completing 14 of 20 trials while contacting the walls in 12 (Section 6.3). System bring-up corroborated the analysis: repeated tuning of yaw rate, trot frequency, and body pitch trim never produced a contact-free in-place turn.

The proposed maneuver replaces the in-place 90$^\circ$ yaw with a discrete three-segment sequence:

\begin{enumerate}
\item A first yaw rotation of 45$^\circ$.
\item A short forward push of approximately 10 cm along the heading achieved after the first segment.
\item A second yaw rotation of 45$^\circ$ in the same direction as the first.
\end{enumerate}

During each yaw segment, a small forward velocity component is retained to prevent the SDK from automatic gait switching; see Section 7.4 for the platform-specific implementation detail.

Geometrically, the sequence converts the swept lateral envelope of a single 90$^\circ$ rotation --- which exceeds the corridor width --- into two smaller swept envelopes connected by a translation segment, each of which fits within the corridor. The forward push translates the body roughly perpendicular to the initial heading, so that after the second 45$^\circ$ rotation the robot is aligned with the new corridor heading and ready to proceed under line-following or open-space behavior. The transition out of the maneuver is triggered by the disappearance of the corridor signature on the lateral range topics.

The three-segment template relaxes the requirement by never demanding the full swept disc: each rotation covers only 45$^\circ$, executed at a stop position where the junction opening absorbs the diagonal excursion. A numerical sweep of the nominal-rectangle envelope over the maneuver (rotation centers and stop position optimized; analysis script available with the study materials) places the feasibility boundary at $c \approx 0.57$ m for this platform --- a reduction of $\approx 0.14$ m ($\approx$20\%) relative to the corner-aware in-place bound. At the deployed operating point ($c = 0.55$ m) the nominal envelope retains a residual interference of $\approx 1.6$ cm at the best stop position, concentrated at the trailing body corner against the outer wall in the mid-maneuver 45$^\circ$ pose; the physical robot nevertheless completed 20 of 20 corridor trials without wall contact (Section 6.3). We read this 1--2 cm discrepancy as the conservatism of the rectangular abstraction --- the platform's rounded, tapering extremities lie inside the nominal bounding box --- and present the computed boundary as a conservative design curve. Figure 5 maps the three regimes over body length and corridor width (in-place feasible / split-turn only / neither), with this course's operating point marked. Kinematically the template is the canonical K-turn of Reeds and Shepp \cite{reeds1990optimal}, which we do not claim as a contribution; the contribution is the feasibility analysis and its hardware corroboration in a regime that prior legged work has addressed only with whole-body MPC \cite{liao2023walking} or morphological compression \cite{buchanan2019walking,buchanan2021perceptive}. The primitive requires no online optimization and no perception beyond the 1-D ranges. Section 6.3 reports the trial outcomes against the in-place-yaw baseline.

\begin{figure*}[t]\centering
\includegraphics[width=0.92\textwidth]{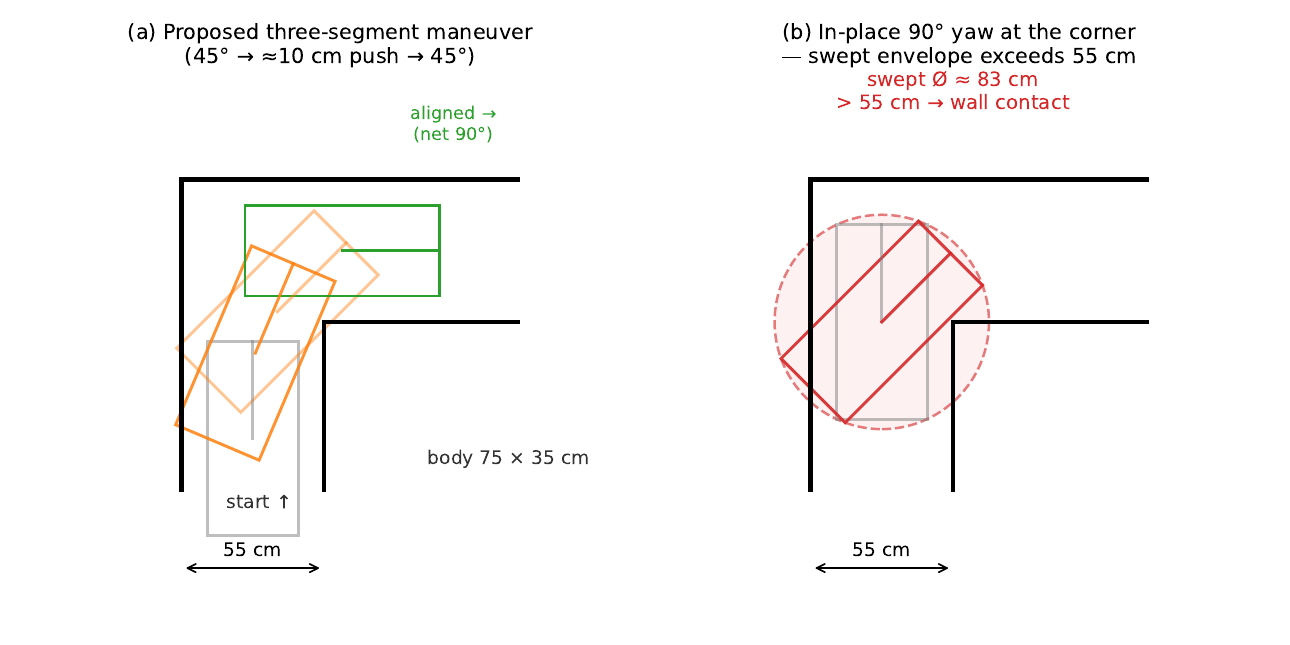}
\caption{\textbf{Narrow-corridor maneuver at a 90$^\circ$ corner.} (a) The proposed three-segment maneuver (45$^\circ$ $\rightarrow$ $\sim$10 cm push $\rightarrow$ 45$^\circ$) threads the 55 cm corridor; (b) an in-place 90$^\circ$ yaw sweeps a circle of diameter $\sim$83 cm $>$ 55 cm and contacts the walls.}
\end{figure*}
\begin{figure*}[t]\centering
\includegraphics[width=0.72\textwidth]{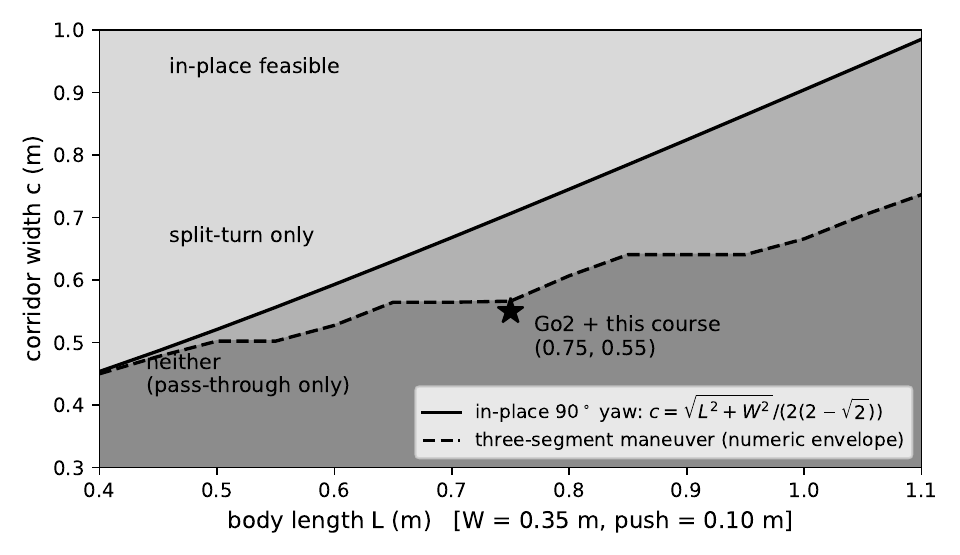}
\caption{\textbf{Geometric feasibility of a 90$^\circ$ corner turn} for a rectangular footprint of length $L$ (width fixed at $W=0.35$ m, push length 0.10 m). Solid curve: corner-aware in-place bound $c=\sqrt{L^2+W^2}/(2(2-\sqrt{2}))$ from the inscribed-disc argument. Dashed curve: numerically computed envelope boundary of the deployed three-segment maneuver (conservative: nominal rectangle, no body rounding). Star: this course ($L=0.75$ m, $c=0.55$ m) --- marginally below the dashed curve yet completed 20/20 trials contact-free, bounding the conservatism of the rectangular abstraction. Regions: in-place feasible / split-turn only / neither (pass-through only).}
\end{figure*}

\subsection{Multi-level stair traversal state machine (\texttt{taijie.cpp v4})}

This subsection instantiates the first contribution of the paper: a fully proprioceptive, sensing-light top-platform-arrival detector for a multi-level stepped platform, whose climb--settle cadence is ablated against continuous climbing in Section 6.4. The course stair structure is a \emph{three-level stepped platform} --- level L1 (90 $\times$ 90 cm), level L2 (70 $\times$ 70 cm), and top level L3 (50 $\times$ 50 cm), separated by 15 cm risers --- rather than a continuous flight. The top platform L3 (50 cm) is smaller than the robot's standing body length ($\approx$75 cm); this is the geometric fact that makes a \emph{timely} summit-arrival decision safety-critical: a decision registered too late lets the robot continue forward, walk off the far edge, and fall (Sections 4.6, 6.4).

The controller is an internal finite-state machine activated by the upper-level coordinator when the front 1D range topic indicates an imminent stair encounter. Its phases are: \texttt{UP} (active upward climbing), \texttt{SUMMIT\_STOP} (settle-and-hold on detected arrival), an optional \texttt{SUMMIT\_BACK} / \texttt{SUMMIT\_STOP2} micro-retreat for re-centering on the small platform, \texttt{SUMMIT\_SHIFT} / \texttt{SUMMIT\_TURN} (the heading change executed on the top platform), \texttt{DOWN} (active descent), and \texttt{EXIT} (return of control, with the gait switched back to the regular static-walk mode). The \texttt{UP  ->  SUMMIT\_STOP} transition --- the detection of top-platform arrival --- is the locus of the contribution and is described next.

\textbf{Climb--settle cadence.} Upward climbing is \emph{not} a continuous trot. The \texttt{UP} phase alternates a short climb burst (forward velocity 0.3 m/s for 1.0 s) with a full stop (2.0 s), repeating until the arrival criterion fires. The motivation is empirical and was confirmed during bring-up: the IMU body-pitch estimate lags and overshoots during continuous climbing, so the pitch sampled \emph{in motion} is a poor estimate of the true body inclination. Stopping lets the estimate settle to a quasi-static value before the arrival criterion is evaluated on it. A stuck-detection guard temporarily raises the climb velocity if planar displacement stalls below 5 cm within a 1.2 s window, and an 18 s watchdog forces the transition if arrival is never declared. (Across the 115 analyzed ablation trials the watchdog never fired; every arrival was detector-declared, the latest at 17.0 s.)

\textbf{How the failure was found.} The overshoot failure first appeared on the deployed integrated stack and was \emph{not} reproducible in a stripped-down stair-only program on the same robot. A controlled comparison traced the difference to the control-loop rate, not the IMU: the integrated stack shares the Jetson with the vision and range pipelines and runs its stair loop at $\approx$15 Hz, while the stair-only program ran at $\approx$50 Hz, where continuous climbing is safe (Section 6.4). This diagnosis fixed the experimental design --- impose the loop period directly and sweep it --- and motivates the dose--response model of Section 4.5: what governs the failure is how far the robot advances per control period.

\textbf{Relative pitch-hysteresis arrival criterion.} During climbing the body pitches nose-up, and the controller continuously tracks the deepest (most negative) settled pitch observed, \texttt{min\_pitch}. Top-platform arrival is declared only when two conditions hold simultaneously, after a minimum climb time of 2 s: (i) a sufficiently deep climb has actually occurred, \texttt{min\_pitch < -24.5}$^\circ$; and (ii) the body has subsequently levelled back out by a margin, \texttt{pitch - min\_pitch > 12}$^\circ$. The key design point is that this is a \emph{relative, two-sided} criterion evaluated on the settled signal, not a single absolute instantaneous threshold. Condition (i) prevents any arrival from being declared before a real climb has been observed; condition (ii) requires a definite return toward level before the summit turn is authorized.

The robustness contribution of this detector is not the arrival rule alone but the \emph{cadence} that precedes it. On the deployed integrated stack the navigation loop shares the onboard compute with the vision and range pipelines and runs at $\approx$15 Hz; at this rate the pitch recovery that signals arrival is registered too late while the robot is still climbing, so a continuous climb can walk off the small top platform before the transition fires. The climb--settle cadence reduces the dependence of the arrival decision on loop rate: during each pause the robot is stationary while the rate-limited loop registers the deep-climb-then-recovery condition, so no additional forward motion is commanded during evaluation. Section 6.4 tests this interpretation by comparing continuous climbing against the climb--settle cadence across controlled loop rates, with the arrival rule held fixed in both arms \cite{johnson2011autonomous,wenger2015semi}.

We note one geometry-specific observation that bounds the mechanism and is reported from the logs in Section 6.4: on this stepped platform the climb produces a \emph{single} continuous nose-up pitch excursion rather than one separable pitch peak per level (the recorded peak-event count at trigger is 1 across all conditions). The arrival criterion therefore relies on the depth-and-recovery of this single excursion, not on counting per-level pitch events; the latter is discussed as a conditional mechanism for staircases with separable per-level transients in Section 7.

\subsection{A latency dose--response model of in-motion summit detection}

The diagnosis of Section 4.4 admits a compact quantitative summary. During continuous climbing the rate-limited loop registers the arrival condition up to a few control periods late, and during each period of delay the robot advances by $v/f$ meters. To first order, overshoot occurs when this latency-induced advance exhausts the geometric margin of the top platform; because the margin and the per-trial latency vary stochastically (gait phase, pitch-estimate transients), we model the overshoot probability of a continuous-climb trial as a logistic dose--response in the \emph{per-period advance}:

\begin{equation}P(\text{overshoot}\mid x) \;=\; \frac{1}{1+e^{-(\alpha+\beta x)}},\qquad x = v/f .\end{equation}

The dose $x$ unifies the two deployment parameters that matter --- climb speed and loop rate --- in one physical quantity, and in particular absorbs the higher climb speed of the $\approx$10 Hz cells (Section 5.3) as a higher dose rather than a confound. We fitted the model in two stages. For a pre-specified held-out check, we first fitted it \emph{before} collecting the 40 Hz and 15 Hz validation cells, by maximum likelihood on the then-available continuous-arm trials ($\approx$30/20/10 Hz): this protocol-clean fit gave $\alpha \approx -0.76$, $\beta \approx 63~\mathrm{m}^{-1}$ ($x^* \approx 0.012$ m per period, a critical loop rate of $\approx$25 Hz at 0.30 m/s), while a sensitivity fit adding the legacy $\approx$50 Hz diagnostic batch (0/10 overshoots; its arrival-margin provenance predates the sweep protocol, Section 6.4) steepened the curve ($\beta \approx 125~\mathrm{m}^{-1}$, $x^* \approx 0.018$ m). The two fits disagree most at 40 Hz --- 43\% versus 22\% predicted overshoot --- a discriminating prediction we specified in writing, together with 80\% bootstrap intervals and the climb--settle prediction ($\leq$10\% per cell at any rate), before collecting the validation cells; the observed 40 Hz cell is consistent with the protocol-clean fit, while the 15 Hz cell falls below both predictions (Section 6.4). For the final reported curve we then refitted on all protocol-clean continuous cells, now including the validation rates ($n = 65$ trials at $\approx$40/30/20/15 Hz, 0.30 m/s, plus the 0.35 m/s $\approx$10 Hz cells): $\alpha \approx -1.03$, $\beta \approx 64~\mathrm{m}^{-1}$, $x^* \approx 0.016$ m per period --- a model-based critical loop rate of $\approx$19 Hz at 0.30 m/s for the tested geometry.

We state the model's scope plainly. It is a two-parameter phenomenological summary fitted at three dose levels, not a mechanistic identification: per-trial forward-odometry drift ($\approx$$\pm$0.1 m of legged-odometry slip on the stairs) exceeds the between-rate differences in latency travel, so the binary outcome --- not the overshoot distance --- is the reliable observable. The operating rule is correspondingly simple within the tested envelope: \emph{keep $v/f$ well below $x^*$, or stop to decide.} During the cadence's decision phase the robot is stationary, making the decision-phase advance approach zero; this explains why the observed cadence-arm overshoot rate is insensitive to the tested loop rates (Section 6.4, Figure 8).

\subsection{Foot-force topology and the dwell component of the arrival check}

The deployed arrival detector of Section 4.4 decides on body pitch alone. Independently of that decision, the robot logs the four \texttt{foot\_force[4]} contact channels throughout every trial. These channels are not consumed by the detector's transition logic in the evaluated stack, but they provide an independent geometric read on whether the \emph{full footprint} has actually settled on the top platform --- a question that matters precisely because the top platform (50 $\times$ 50 cm) is smaller than the robot's standing length ($\approx$75 cm), leaving a front/back overhang of a few centimetres when the robot is centered. Had a summit turn been committed before all four feet were supported, the 90$^\circ$ yaw could have pivoted a foot over the platform edge, with consequences ranging from loss of heading control to a fall.

We therefore report the longest interval during which all four foot-force channels are simultaneously above a contact threshold. This four-foot dwell is one component of the overshoot criterion (Section 5.4) and is computed from a channel the pitch-based detector does not use. On the climb--settle runs it is long and stable ($\approx$4 s for the representative trial), consistent with the hold landing the full footprint on the platform; on the continuous-overshoot trials it is short or absent (below 0.1 s), consistent with the robot tipping over the far edge rather than settling. Because the dwell enters the overshoot label, we do \emph{not} present it as independent evidence of overshoot; it is, however, independent of the pitch-based \emph{control decision}, and we report it to show directly that the cadence lands a stable four-foot stance while continuous overshoot does not. The long dwell is also partly a mechanical consequence of the hold --- the robot stops on the platform by design --- a further reason to read it as descriptive rather than as proof of correct arrival. This use of the \emph{simultaneous four-leg force topology} as a discrete platform-occupancy read is distinct from the literature, which uses per-leg foot-force or touchdown timing as a gait-phase signal \cite{han2023contact} or a terrain-type classifier \cite{wang2023climbing}.

For full provenance we record one implementation history. An earlier exploratory variant of the detector additionally used the all-four-feet dwell as an explicit confirmation \emph{gate} on the \texttt{UP  ->  SUMMIT} transition. We found during field bring-up that the pitch-hysteresis-with-settling criterion alone was more robust for competition deployment, and the evaluated detector relies on it; the foot-force confirmation gate is retained as a future-work direction (Section 7) for deployments where the small-platform geometry constraint must be enforced \emph{in the control loop} rather than verified post hoc. The results of Section 6.4 and the reference implementation correspond to the deployed pitch-hysteresis detector.

Figures accompanying Section 4: Figure 4 (corridor geometry and three-segment maneuver schematic), Figure 5 (corner-turn feasibility chart), and Figure 6 (stair internal state machine). The stair-ablation and dose--response figures are presented with the results in Section 6.4.
\begin{figure*}[t]\centering
\includegraphics[width=0.92\textwidth]{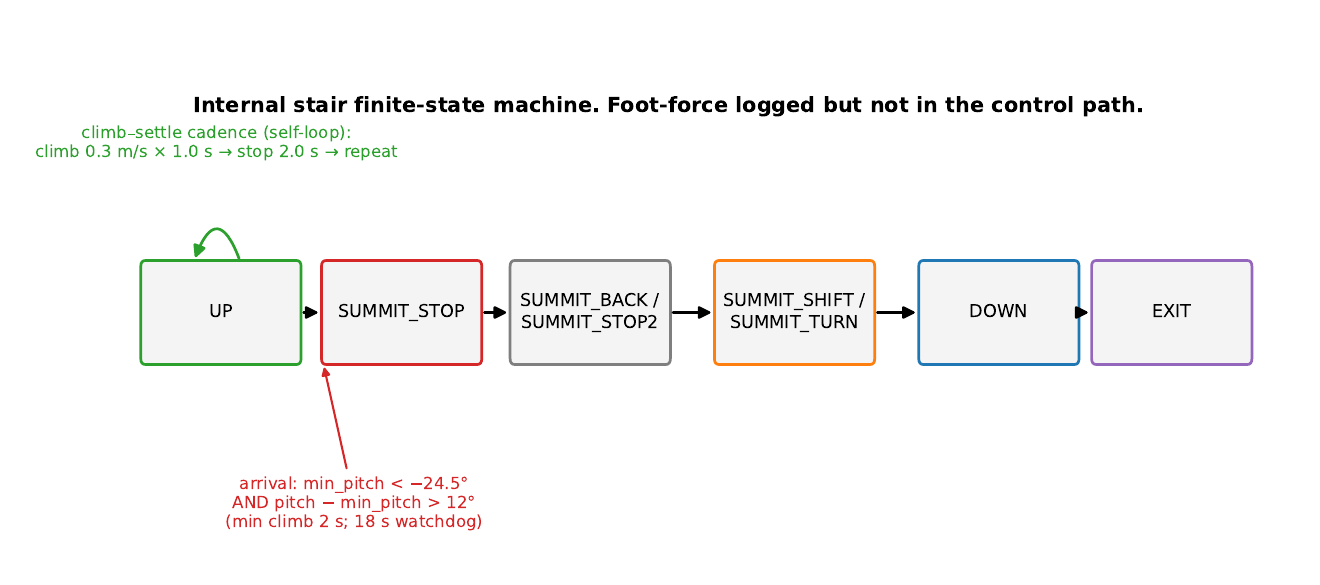}
\caption{\textbf{Internal stair finite-state machine} (\texttt{taijie.cpp v4}). The highlighted \texttt{UP -> SUMMIT\_STOP} transition instantiates the proposed detector: under the climb--settle cadence, top-platform arrival is declared on a relative pitch-hysteresis criterion (min pitch $< -24.5^\circ$, then pitch $-$ min $> 12^\circ$). Foot force is logged but does not gate any transition.}
\end{figure*}

\section{Experimental Setup}

This section documents the physical course, the robot configuration used for the trials, the trial protocol, and the metrics reported in Section 6. The intent is full reproducibility: a reader with access to a Unitree Go2 EDU and a structurally similar course should be able to replicate the entire procedure from this section together with Sections 3 and 4.

\subsection{Robot and software configuration}

All trials are run on the Unitree Go2 EDU robot described in Section 3.1, with the NVIDIA Jetson Orin as the onboard compute platform. The operating system recorded during experiment documentation is Ubuntu 20.04.5 LTS (\texttt{focal}). The local robot environment did not expose a ROS 2 runtime: \texttt{ros2 --version} returned \texttt{command not found}, and \texttt{\$ROS\_DISTRO} was unset. Accordingly, this manuscript describes the experimental stack as a C++ / \texttt{unitree\_sdk2} DDS implementation rather than a ROS 2 Foxy implementation. The \texttt{unitree\_sdk2} interface was installed from Unitree's public SDK2 repository, \url{https://github.com/unitreerobotics/unitree_sdk2}; the exact local commit hash was not recorded, so exact-version reproducibility is bounded to the public repository rather than to a pinned commit. The active controller implementation uses the line-following, obstacle/corridor, stair, platform, warning-action, and terminal routines coordinated by the integrated C++ program. The source is available from the corresponding author on reasonable request and is intended for public release after repository cleanup. For provenance, the integrated stack was originally developed for and deployed in a national robotics competition (RAICOM); the reference implementation is the single integrated program described in Sections 3--4, and this paper reports a controlled evaluation of that deployed system rather than a competition account.

The Go2 is operated in its standard trot gait throughout all trials, except where Section 4.3 requires the small forward velocity \texttt{vx = 0.02} to suppress automatic SDK gait switching. No tether other than a passive safety leash is used; control commands and sensor data flow entirely between the Jetson Orin and the Go2 base controller through the on-robot communication bus. The experiment checklist did not record a fixed battery-percentage cutoff, so battery effects are not analyzed as an independent experimental variable in this manuscript.

Table 2 lists the platform configuration and the key control constants referenced throughout Sections 3--4; the line-following and open-space-avoidance gains (PD coefficients, region-of-interest bounds, range trigger thresholds) are standard and are provided with the reference implementation rather than reproduced here.

\begin{table}[t]\centering
\caption{\textbf{System configuration and key control parameters.}}
\footnotesize\begin{tabularx}{\columnwidth}{@{}>{\raggedright\arraybackslash}p{1.3cm} >{\raggedright\arraybackslash}p{2.9cm} >{\raggedright\arraybackslash}X@{}}
\toprule
Group & Parameter & Value \\
\midrule
Platform & robot / onboard compute & Unitree Go2 EDU / NVIDIA Jetson Orin \\
Platform & OS / interface & Ubuntu 20.04.5 LTS / \texttt{unitree\_sdk2} (C++ DDS), no ROS 2 \\
Platform & gait & standard trot \\
Sensing & modalities used & IMU; 4$\times$ foot-force; 3$\times$ 1D range (\texttt{rt/utlidar/\allowbreak range\_info}); monocular camera (line extraction only) \\
Corridor maneuver & segment sequence & 45$^\circ$ yaw $\rightarrow$ $\approx$ 10 cm push $\rightarrow$ 45$^\circ$ yaw \\
Corridor maneuver & yaw forward velocity & \texttt{vx} = 0.02 m/s (suppresses SDK gait switching) \\
Stair detector & climb velocity (\texttt{VX\_UP}) & 0.3 m/s \\
Stair detector & climb burst / stop (cadence) & 1.0 s / 2.0 s \\
Stair detector & deep-climb threshold (\texttt{PITCH\_DEEP}) & min pitch $< -24.5^\circ$ \\
Stair detector & recovery margin (\texttt{PITCH\_RECOVERY}) & pitch $-$ min pitch $> 12^\circ$ (deployed); $16^\circ$ in the ablation campaign (Section 5.3) \\
Stair detector & minimum climb time & 2.0 s \\
Stair detector & stuck guard & $< 5$ cm planar displacement in 1.2 s $\rightarrow$ raise climb velocity \\
Stair detector & arrival watchdog & 18 s \\
Foot-force & contact threshold & 24 sensor units \\
\bottomrule
\end{tabularx}
\end{table}

\subsection{Physical course layout}

The course is laid out indoors on a flat tiled floor and consists of three segments traversed in sequence:

\begin{enumerate}
\item \textbf{Line-following segment.} A black tape track with a measured width of 100 mm is laid down on the floor. The total track length is approximately 15 m by field estimate and includes 12 heading changes, each nominally 90 deg. The line-following segment includes a deliberate 3 cm gap and one occluded segment where the line is briefly hidden under a baffle, in order to exercise the lost-line recovery sub-state of Section 4.1.

\item \textbf{Obstacle / corridor segment.} The track passes through a region in which two parallel rigid walls are installed at a separation of 55 cm. The corridor width was measured as 550 mm at the entrance, middle, and exit, so the minimum recorded width is 55 cm. The walls extend for a length of approximately 200 cm. Outside the corridor segment, but along the same general region, a single rigid obstacle is placed approximately on the line so that the robot must perform an open-space arc avoidance before re-acquiring the line. The corridor entry and exit points are unmarked: the upper-level coordinator detects the corridor geometry from the lateral 1D range topics as described in Section 4.2.

\item \textbf{Multi-level staircase segment.} The staircase consists of three levels separated by intermediate platforms, terminating in a small top platform of dimensions 50 x 50 cm. Each riser is approximately 15 cm and each tread is approximately 10 cm. The intermediate platform footprints are 90 x 90 cm at L1 and 70 x 70 cm at L2; the L3 top platform is 50 x 50 cm. The three 15 cm risers give cumulative platform elevations of 15 cm, 30 cm, and 45 cm above the floor; we report these as the as-built dimensions of the constructed platform. The top platform is intentionally smaller than the robot's measured standing length of approximately 75 cm; measured at the foot contacts (the front-to-rear foot span is shorter than the body length), this gives a front/back overhang of approximately 7 cm when the robot is centered.
\end{enumerate}

Table 3 summarizes the course geometry and the per-segment task metrics; Figure 1 in Section 3 is a schematic plan of the assembled course with the three segments labeled.

\begin{table}[t]\centering
\caption{\textbf{Course geometry and task metrics} (as-built dimensions of the physical course).}
\footnotesize\begin{tabularx}{\columnwidth}{@{}>{\raggedright\arraybackslash}p{1.9cm} >{\raggedright\arraybackslash}p{2.5cm} >{\raggedright\arraybackslash}X@{}}
\toprule
Segment & Parameter & Value \\
\midrule
Line-following & painted-tape width & 100 mm \\
Line-following & track length (field estimate) & $\approx$ 15 m \\
Line-following & heading changes & 12 $\times$ $\approx$ 90$^\circ$ \\
Line-following & deliberate discontinuities & 3 cm gap; 1 baffle occlusion \\
Narrow corridor & corridor width & 550 mm \\
Narrow corridor & wall length & $\approx$ 200 cm \\
Multi-level platform & level footprints (L1 / L2 / L3) & 90$\times$90 / 70$\times$70 / 50$\times$50 cm \\
Multi-level platform & riser / tread & $\approx$ 15 cm / $\approx$ 10 cm \\
Multi-level platform & cumulative elevations (L1 / L2 / L3) & 15 / 30 / 45 cm \\
Robot & standing body (length $\times$ width) & $\approx$ 75 $\times$ 35 cm \\
Robot & top-platform overhang (at foot contacts) & $\approx$ 7 cm \\
\bottomrule
\end{tabularx}
\end{table}

\subsection{Trial matrix}

The trial campaign comprises four categories of formal run, summarized below and elaborated in Table 4.

\textbf{End-to-end integrated runs.} Twenty trials of the complete course are run, starting from a standing pose at the beginning of the line-following segment and terminating at the \texttt{EXIT} state of the stair traversal sub-controller. Each trial is logged for success/failure, completion time, the number of upper-level state transitions, and the failure mode if applicable (intervention, fall, abort).

\textbf{Per-module verification runs.} The line-following sub-controller is exercised in 10 isolation trials on the line segment without obstacles. A 40-trial corridor campaign is run separately as described below. Open-space avoidance is exercised as part of the integrated route but is not reported as an independent 10-trial isolation campaign.

\textbf{Narrow-corridor maneuver campaign.} Twenty trials of the corridor segment are run with the three-segment maneuver of Section 4.3 active, paired with 20 trials in which the upper-level coordinator is configured to attempt an in-place 90 deg yaw instead. The latter is the geometric-infeasibility baseline of Section 6.3.

\textbf{Stair detector ablation.} This is the central ablation of the paper and follows the protocol summarized in Table 4. The deployed detector pairs a climb--settle cadence with a relative pitch-hysteresis arrival criterion (Section 4.4); the ablation isolates the \emph{cadence} by comparing continuous climbing against climb--settle while holding the arrival rule and all constants fixed. Because the overshoot failure is driven by control-loop latency (Section 6.4), the cadence factor is crossed with a controlled control-loop rate spanning $\approx$40, $\approx$30, $\approx$20, and $\approx$15 Hz, imposed through the loop period; the $\approx$15 Hz cells sit at the deployed integrated-stack rate, and the $\approx$40 Hz cells provide the out-of-sample model test of Section 4.5 (continuous arm only). The climb speed (0.30 m/s) and the arrival constants are fixed within each arm. One protocol detail is disclosed for exactness: while the deployed competition detector uses a recovery margin of 12$^\circ$ (Table 2), the entire ablation campaign was run with a deliberately later-triggering margin of 16$^\circ$, applied equally to both arms --- the cadence contrast is therefore internally consistent, and the deployed-margin behaviour at the deployed rate is documented separately by the integrated end-to-end runs (Section 6.1).

\begin{table}[t]\centering
\caption{\textbf{Stair detector ablation protocol.} The deployed climb--settle detector of Section 4.4 is compared against continuous climbing (cadence disabled, same arrival rule) across four imposed control-loop rates at a uniform 0.30 m/s climb speed, with a legacy 0.35 m/s $\approx$10 Hz pair retained as a sensitivity condition. The $\approx$40 Hz continuous cell provides the out-of-sample model test (Section 4.5).}
\footnotesize\begin{tabularx}{\columnwidth}{@{}>{\raggedright\arraybackslash}X >{\raggedright\arraybackslash}p{2.5cm} r@{}}
\toprule
Control-loop rate & Cadence & Trials \\
\midrule
$\approx$40 Hz & Continuous climb & 15 \\
$\approx$30 Hz & Continuous climb & 15 \\
$\approx$30 Hz & Climb--settle (proposed) & 15 \\
$\approx$20 Hz & Continuous climb & 15 \\
$\approx$20 Hz & Climb--settle (proposed) & 15 \\
$\approx$15 Hz & Continuous climb & 15 \\
$\approx$15 Hz & Climb--settle (proposed) & 15 \\
$\approx$10 Hz (0.35 m/s, sensitivity) & Continuous climb & 5 \\
$\approx$10 Hz (0.35 m/s, sensitivity) & Climb--settle (proposed) & 5 \\
\bottomrule
\end{tabularx}
\end{table}

Each protocol cell contains n = 15 trials at a uniform climb speed of 0.30 m/s, eliminating the speed difference the earlier pilot sweep carried; the legacy $\approx$10 Hz cells (run at 0.35 m/s, n = 5) are retained only as a higher-dose sensitivity condition, their speed difference absorbed by the dose--response model of Section 4.5 (dose x = v/f). A distinct \emph{non-arrival} (premature) outcome --- the robot triggering before the full footprint reached the top platform, so that no overshoot opportunity arose --- is recorded separately from overshoot and reported in Section 6.4.

\subsection{Metrics}

The following metrics are reported in Section 6.

\textbf{Success rate} is the binary success/failure per trial under the failure conditions defined for that trial category (intervention, fall, summit overshoot, abort), and \textbf{completion time} is the interval from activation of the relevant sub-controller to its successful termination, in seconds. \textbf{Recovery latency} (line-following) is the time from loss of the painted line to its re-acquisition under the lost-line recovery sub-state, where applicable. The \textbf{number of state-machine interventions} counts upper-level coordinator transitions during a trial as a proxy for trial complexity, and \textbf{falls and aborts} are counted separately as raw counts --- a fall being operationally defined as loss of stable standing pose, body contact with the ground or course furniture, or any instability requiring immediate operator stop.

Three metrics are specific to the stair ablation. The \textbf{summit-arrival overshoot} label classifies each stair trial as \emph{overshoot} or \emph{safe arrival} from the logged traces: an overshoot is a nose-down body-pitch excursion past +12$^\circ$ (the robot tipping over the far edge) together with a longest all-four-foot dwell below 1.0 s (no stable summit stance). This criterion is applied uniformly to every trial and agrees with operator observation except at a small number of marginal or operator-assisted trials (Section 7.4); an independent coder, blind to the trace data and to the experimental condition, additionally labeled every newly recorded trial from video, with agreement reported as Cohen's $\kappa$ (Section 7.4). The \textbf{non-arrival (premature) outcome} is a trial in which the detector declared summit arrival while the robot's full footprint had not reached the top platform (one or more rear feet remaining on a lower step); because the foot-force channels register contact on any step and cannot by themselves distinguish the top platform from a lower one, it is established by visual observation rather than from the trace alone --- recorded by the operator at the trial and independently reproduced by the blind video coder (Section 7.4) --- and is mutually exclusive with overshoot. Finally, the \textbf{forward-displacement overshoot distance} (descriptive only) is the forward odometry displacement logged through the ascent; because legged-odometry drift on the stairs ($\approx$$\pm$0.1 m, with an inconsistent sign across batches) exceeds the between-rate latency-travel differences, it is reported descriptively and is not used as a model observable, the binary overshoot outcome being the reliable observable (Section 4.5).

\textbf{Statistical plan.} Rates are reported with Wilson 95\% confidence intervals. Confirmatory comparisons are pre-specified as follows. (H1) Under continuous climbing, overshoot probability increases with the per-period advance $x=v/f$ (logistic dose--response; Section 4.5). (H2) At each loop rate, the climb--settle cadence reduces overshoot relative to continuous climbing: per-rate Fisher exact tests, with a rate-stratified Cochran--Mantel--Haenszel test and the Mantel--Haenszel common odds ratio as the aggregate estimate. (H3) The dose--response model predicts the overshoot rate at two loop rates never used in fitting (40 Hz and 15 Hz; both predictions were specified in writing on 2026-06-10, before the validation data were collected, and both are reported in Section 6.4). Corridor comparisons use Fisher exact tests for failure and wall-contact rates and Mann--Whitney U tests for exit-heading error and traversal time. All statistics are computed from the raw traces by the analysis scripts available with the study materials.

\subsection{Reset procedure between trials}

After each trial, the robot is returned to the starting pose at the head of the line-following segment by the operator. Course furniture (corridor walls, obstacle, stair configuration) is verified against the course schematic and adjusted if displacement has occurred. The robot is rebooted between trial categories to avoid carrying over residual state in the SDK.

\section{Results}

All continuous metrics in this section were re-derived directly from the per-trial raw onboard trace files (body pitch, four-channel foot force, 1D range, and end-to-end and line-following timing) by an analysis script, rather than read from a summary trial log. Success and platform-outcome labels (L1/L2/L3, operator intervention) are taken from the onboard detector log together with operator annotation at trial time, because they are recorded observations of controller behaviour. The independent blind video coding of Section 7.4 covers the newly recorded stair-ablation trials; these system-level success labels carry no such independent check, a residual caveat for Sections 6.1--6.3. Standard deviations are sample standard deviations (n - 1).

This section reports the verified physical-trial records from the experimental campaign. Results are given as raw counts, percentages, and descriptive statistics, accompanied by the pre-specified confirmatory tests of Section 5.4 (per-rate and pooled Fisher exact, rate-stratified Cochran--Mantel--Haenszel, and Mann--Whitney U) for the stair ablation and corridor comparisons. Each reported rate carries a Wilson score 95\% confidence interval for the underlying binomial proportion.

The campaign contains 185 analyzed trials: 115 stair-detector ablation trials (117 were executed; the two continuous trials whose per-iteration traces failed to record are excluded throughout, Section 6.4), 40 narrow-corridor trials, 20 end-to-end integrated-course trials, and 10 line-following isolation trials. Unless otherwise stated, timing is measured from activation of the relevant controller to its completion or failure event. Two metric definitions are used throughout the corridor analysis: the \emph{exit heading error} is the absolute deviation between the robot heading and the corridor axis at the moment the robot clears the corridor, and the \emph{peak heading deviation} is the maximum such absolute deviation observed during the maneuver. The \emph{top-platform foot-force dwell} is the longest interval during which all four foot-force readings are simultaneously above a contact threshold of 24 (sensor units); the threshold is an operational definition taken from the suspended-versus-standing self-check ($\approx$0 with the feet off the ground, above $\approx$28 when standing), and the qualitative dwell contrast reported below is insensitive to its exact value.

\subsection{End-to-End Course Performance}

Across 20 end-to-end trials of the complete inspection-style course, the integrated stack completed 18 trials, a 90.0\% success rate (Wilson 95\% CI [69.9, 97.2]\%). Mean completion time was 77.30 $\pm$ 4.40 s, with a minimum of 70.70 s and a maximum of 85.57 s. The two failed trials did not arise from the stair detector or the corridor maneuver: one failed because of late line reacquisition near the final obstacle, and one required operator intervention after platform drift.

\begin{table}[t]\centering
\caption{\textbf{End-to-end integrated course results.}}
\footnotesize\begin{tabular}{@{}>{\raggedright\arraybackslash}p{5.0cm}l@{}}
\toprule
Metric & Value \\
\midrule
Total trials & 20 \\
Successful trials & 18 \\
Success rate & 90.0\% \\
Mean completion time & 77.30 $\pm$ 4.40 s \\
Min / max completion time & 70.70 / 85.57 s \\
Operator-intervention trials & 1 \\
Wall-contact trials & 0 \\
Line-following failure / late reacquisition & 1 \\
Platform drift requiring intervention & 1 \\
\bottomrule
\end{tabular}
\end{table}

These results support the system-level claim that the deterministic sub-controllers can be chained under a single state-machine stack to complete the structured inspection route. The residual failures also locate the remaining system risk: late-stage line reacquisition and platform-level alignment, not the proposed stair confirmation mechanism.

\subsection{Line-Following Performance}

The line-following module was tested in 10 isolation trials, all of which completed successfully (100.0\% success). Mean traversal time was 30.04 $\pm$ 0.49 s. Lost-line recovery was invoked in 2 of the 10 trials (mean 0.20 $\pm$ 0.42 events per trial); the maximum recovery latency observed across the campaign was 0.85 s, and the mean per-trial maximum recovery time was 0.16 $\pm$ 0.34 s.

\begin{table}[t]\centering
\caption{\textbf{Line-following isolation results.}}
\footnotesize\begin{tabular}{@{}>{\raggedright\arraybackslash}p{5.0cm}l@{}}
\toprule
Metric & Value \\
\midrule
Total trials & 10 \\
Successful trials & 10 \\
Success rate & 100.0\% \\
Mean traversal time & 30.04 $\pm$ 0.49 s \\
Trials with lost-line events & 2 \\
Mean lost-line events per trial & 0.20 $\pm$ 0.42 \\
Mean max recovery time per trial & 0.16 $\pm$ 0.34 s \\
Maximum observed recovery time & 0.85 s \\
\bottomrule
\end{tabular}
\end{table}

These data are treated as supporting evidence rather than a primary contribution. The line follower provides a stable route-following primitive for the integrated route, but its ROI-thresholding and PD-control structure remain classical.

\subsection{Narrow-Corridor Maneuver Performance}

The corridor experiment compared the proposed split-turn maneuver against an in-place-yaw baseline across 20 trials per condition in a 550 mm corridor. The split-turn maneuver completed all 20 trials without wall contact (100.0\%, Wilson 95\% CI [83.9, 100.0]\%), while the in-place-yaw baseline completed 14 of 20 trials (70.0\%, [48.1, 85.5]\%) and made wall contact in 12 of 20; 6 of those contacts caused trial failure. The split-turn maneuver was also faster (18.08 $\pm$ 0.89 s versus 24.08 $\pm$ 0.89 s).

Heading quality separates the two conditions most clearly. The split-turn maneuver exited the corridor with a mean exit heading error of 1.56 $\pm$ 0.71$^\circ$ (range 0.03--2.53$^\circ$), whereas the in-place-yaw baseline exited at 5.64 $\pm$ 2.50$^\circ$ (range 1.32--8.59$^\circ$): the baseline is both less aligned and far less consistent at exit. The peak heading deviation during the maneuver was tightly clustered for each condition (split-turn 2.54 $\pm$ 0.01$^\circ$, in-place-yaw 8.82 $\pm$ 0.02$^\circ$), reflecting the fixed geometry of each maneuver. The split-turn maneuver maintained a minimum lateral clearance of 0.265 m throughout every trial, while the in-place-yaw baseline dropped to as little as 0.057 m, consistent with its wall-contact count.

\begin{table}[t]\centering
\caption{\textbf{Narrow-corridor maneuver outcomes.}}
\footnotesize\begin{tabularx}{\columnwidth}{@{}>{\raggedright\arraybackslash}p{3.0cm} >{\raggedright\arraybackslash}X >{\raggedright\arraybackslash}X@{}}
\toprule
Outcome & Split-turn maneuver & In-place-yaw baseline \\
\midrule
Total trials & 20 & 20 \\
Successful trials & 20 & 14 \\
Success rate & 100.0\% & 70.0\% \\
Failed trials & 0 & 6 \\
Wall-contact trials & 0 & 12 \\
Mean corridor time & 18.08 $\pm$ 0.89 s & 24.08 $\pm$ 0.89 s \\
Min / max corridor time & 16.66 / 19.70 s & 22.49 / 25.43 s \\
Mean exit heading error & 1.56 $\pm$ 0.71$^\circ$ & 5.64 $\pm$ 2.50$^\circ$ \\
Peak heading deviation & 2.54 $\pm$ 0.01$^\circ$ & 8.82 $\pm$ 0.02$^\circ$ \\
Minimum lateral range (side 1-D sensor) & 0.265 m & 0.057 m \\
\bottomrule
\end{tabularx}
\end{table}

The corridor results support the mechanism-level argument for the discrete maneuver. The baseline could sometimes complete the corridor, but even successful baseline trials often involved wall contact, whereas the split-turn primitive avoided contact across all trials and reduced both traversal time and exit heading error. The claim remains bounded: the maneuver is a structured-course primitive for a known corridor-width regime, not a general narrow-passage planner.

\subsection{Stair Detector Ablation: Robustness to Control-Loop Latency}

The deployed stair-summit detector pairs a climb--settle cadence with a relative pitch-hysteresis arrival criterion (Section 4.4). The ablation reported here isolates the \textbf{cadence}: it compares continuous climbing (cadence disabled) against the climb--settle cadence, holding the arrival rule and every other constant fixed, so that the only manipulated variable is whether the robot pauses to let the body-pitch estimate settle before the arrival condition is evaluated.

\textbf{Loop-rate protocol.} Following the diagnosis of Section 4.4, the ablation sweeps the imposed control-loop rate across $\approx$40, $\approx$30, $\approx$20, and $\approx$15 Hz --- the $\approx$15 Hz cells at the deployed integrated-stack rate and the $\approx$40 Hz cells as an out-of-sample model test --- by adding a fixed per-iteration delay that throttles the loop to the target period. This reproduces the reduced update rate but not the timing jitter of true compute contention, a caveat for reading the deployed-rate comparison. At $\approx$50 Hz, the stair-only program's natural rate, continuous climbing is safe: the legacy diagnostic batch at that rate records 0/10 overshoots under continuous climbing and 0/11 under climb--settle (trace-derived with the same criterion); it is excluded from the fitted matrix because its arrival-margin provenance predates the sweep protocol, and enters only a sensitivity variant of the model fit (Section 4.5).

\textbf{Overshoot criterion.} Because the top platform L3 (50 $\times$ 50 cm) is shorter than the robot, an arrival registered too late lets the robot continue forward, walk off the far edge, and tip nose-down. We therefore classify a stair trial as an \textbf{overshoot} when the body-pitch trace shows a nose-down excursion past +12$^\circ$ (the body tipping over the edge) \emph{and} the longest all-four-foot dwell is below 1.0 s (no stable summit stance). This criterion is applied uniformly to every logged trial from the pitch and foot-force traces; it agrees with operator observation at the trial and was cross-checked against an independent blind video coding of the newly recorded trials (Cohen's $\kappa$; Section 7.4).

As a transparency check on the label itself, we note its sensitivity to the dwell conjunct. Dropping the dwell term and classifying on the pitch excursion alone (nose-down past +12$^\circ$) makes the criterion strictly more inclusive, so each arm's overshoot count can only stay equal or rise --- a pitch-only label is therefore \emph{less} favourable to the cadence, not more. The dwell term does its discriminating work mainly in the climb--settle arm, where a trial may tip briefly during the climb transient yet then settle into a stable four-foot stance --- a non-overshoot under the two-part label that a pitch-only label would instead count as an overshoot --- whereas continuous-overshoot trials walk off the platform without ever settling and are largely unaffected by the choice. We retain the two-part label because the dwell term encodes the operationally relevant outcome --- ending in a stable four-foot stance on the short top platform --- that a pitch threshold alone does not capture. Recomputing the label on the pitch excursion alone across the full trace set bears this out: it raises the $\approx$30 Hz climb--settle count from 1/15 to 2/15 and the $\approx$10 Hz sensitivity cell from 0/5 to 1/5 (the $\approx$20 and $\approx$15 Hz climb--settle cells stay at 0/15). On the continuous side a single $\approx$15 Hz trial moves the other way --- an operator caught the robot after it had already tipped over the edge, and the resulting manual hold produced a stable dwell that the two-part label scored as a safe arrival, which the pitch-only label restores to an overshoot (7/15 $\rightarrow$ 8/15). Both adjustments leave the climb--settle arm far below the continuous arm under either label, so the dwell conjunct does not manufacture the cadence effect; we flag the single operator-assisted trial among the limitations (Section 7.4).

\begin{table*}[t]\centering
\caption{\textbf{Stair-detector overshoot ablation: summit-arrival overshoot rate by control-loop rate and cadence.} Overshoot is defined in the text (a nose-down body-pitch excursion past +12$^\circ$ together with a longest all-four-foot dwell below 1.0 s). Brackets give Wilson 95\% confidence intervals. The deployed integrated stack runs at $\approx$15 Hz. Labels are re-derived from the raw per-trial pitch and foot-force traces by the analysis script; two continuous trials whose per-iteration traces failed to record are omitted, since the trace-based criterion is undefined for them (the same rule applied to both arms).}
\footnotesize\begin{tabular}{@{}>{\raggedright\arraybackslash}p{2.7cm}llll@{}}
\toprule
Control-loop rate & Cadence & Trials & Overshoot & Overshoot rate (95\% CI) \\
\midrule
$\approx$40 Hz & Continuous climb & 15 & 5 & 33\% [15, 58]\% \\
$\approx$30 Hz & Continuous climb & 15 & 6 & 40\% [20, 64]\% \\
$\approx$30 Hz & Climb--settle (proposed) & 15 & 1 & 7\% [1, 30]\% \\
$\approx$20 Hz & Continuous climb & 15 & 9 & 60\% [36, 80]\% \\
$\approx$20 Hz & Climb--settle (proposed) & 15 & 0 & 0\% [0, 20]\% \\
$\approx$15 Hz & Continuous climb & 15 & 7 & 47\% [25, 70]\% \\
$\approx$15 Hz & Climb--settle (proposed) & 15 & 0 & 0\% [0, 20]\% \\
$\approx$10 Hz (sensitivity, 0.35 m/s) & Continuous climb & 5 & 4 & 80\% [38, 96]\% \\
$\approx$10 Hz (sensitivity, 0.35 m/s) & Climb--settle (proposed) & 5 & 0 & 0\% [0, 43]\% \\
\bottomrule
\end{tabular}
\end{table*}

Under continuous climbing the overshoot rate rises as the per-period advance grows --- 5/15 (33\%), 6/15 (40\%), and 9/15 (60\%) at $\approx$40, $\approx$30, and $\approx$20 Hz --- whereas the climb--settle cadence holds it near zero at every rate --- 1/15 (7\%) at $\approx$30 Hz and 0/15 at $\approx$20 and $\approx$15 Hz (Table 8, Figure 7). At the deployed $\approx$15 Hz operating point the continuous arm overshoots in 7/15 trials against 0/15 for the cadence: the cadence is the difference between routine overshoot and routine safe arrival. We report Wilson 95\% confidence intervals per cell and the pre-specified aggregate tests of Section 5.4 (H2): pooled over the matched-speed $\approx$30/20/15 Hz matrix, continuous climbing overshoots in 22/45 trials versus 1/45 under the cadence (Fisher exact p $\approx$ $2.4 \times 10^{-7}$); the rate-stratified CMH test gives $\chi^{2}$(1) = 22.7, p $\approx$ $1.9 \times 10^{-6}$, with a Mantel--Haenszel common odds ratio of $\approx$36 (95\% CI [4.8, 268]). Per-rate Fisher tests are significant at $\approx$20 Hz (p $\approx$ $7 \times 10^{-4}$) and $\approx$15 Hz (p $\approx$ $6 \times 10^{-3}$); at $\approx$30 Hz the single cadence-arm overshoot widens the interval (p $\approx$ 0.08), so the per-cell tests are read together with the pooled and stratified results (Section 7.4). The 0.35 m/s $\approx$10 Hz sensitivity cells (continuous 4/5, cadence 0/5; Fisher p $\approx$ 0.048) are consistent with the matrix and are folded into the dose--response model of Section 4.5 as the highest-dose condition, their speed difference carried by the dose x = v/f rather than treated as an across-rate confound. The design is fully crossed; the two continuous trials omitted from the trace-based labelling --- one in the $\approx$30~Hz cell and one in the $\approx$10~Hz sensitivity cell --- recorded empty per-iteration trace files, so the trace-based criterion is undefined for them; they were not re-run, and the per-cell $n$ already excludes them (the $\approx$30~Hz continuous cell drew 16 trials for 15 valid traces, the $\approx$10~Hz sensitivity cell 6 for 5), the same objective rule applied to both arms (Section 7.4).
\begin{figure}[t]\centering
\includegraphics[width=\linewidth]{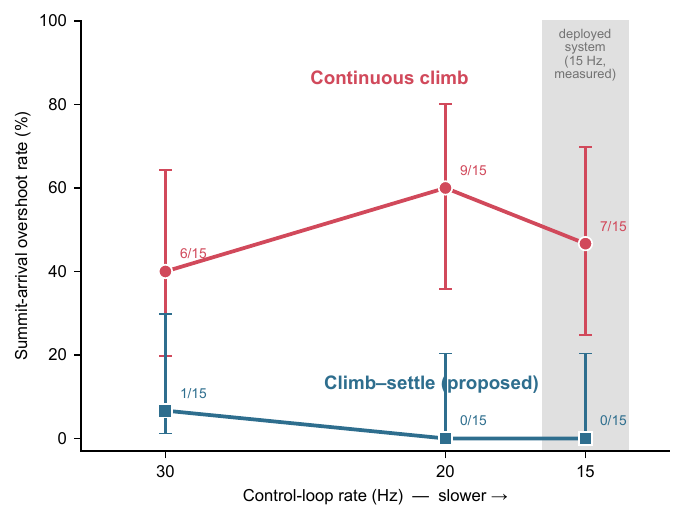}
\caption{\textbf{Stair-detector ablation:} summit-arrival overshoot rate versus control-loop rate, continuous climbing versus the climb--settle cadence. Overshoot is a nose-down body-pitch excursion past +12$^\circ$ together with a longest all-four-foot dwell below 1.0 s. Error bars are Wilson 95\% confidence intervals; per-cell trial counts ($n = 15$) are shown. The deployed integrated stack runs at $\sim$15 Hz. Continuous detection overshoots more as the loop slows, whereas the cadence holds overshoot near zero.}
\end{figure}

\textbf{Dose--response (H1).} Treating each continuous-arm trial as a binary outcome at its per-period advance x = v/f collapses the rate sweep onto a single axis (Figure 8). The final logistic fit over all protocol-clean continuous cells (n = 65) places the 50\% dose at x* $\approx$ 0.016 m per period --- a critical loop rate of $\approx$19 Hz at 0.30 m/s --- while the climb--settle arm stays at 0--7\% across the same dose range, as the decoupling mechanism predicts.

\textbf{Pre-specified out-of-sample checks (H3).} The two dose--response fits of Section 4.5, specified in writing on 2026-06-10 before the validation cells were collected, made point predictions at \emph{two} held-out rates; we report both as pre-specified. At 40 Hz the fits differed most --- 43\% (protocol-clean) versus 22\% (the steeper sensitivity fit including the legacy $\approx$50 Hz batch) --- and the observed cell overshot in 5/15 trials (33\%, Wilson 95\% CI [15, 58]): this lies inside the 80\% pre-specified interval of the protocol-clean fit ([18, 63]) and just above that of the sensitivity fit ([9, 31]), so the 40 Hz observation is consistent with the protocol-clean fit and in mild tension with the steeper one. We are deliberately measured about its strength: the observed 33\% is almost equidistant from the two point predictions, the discrimination rests on the pre-specified 80\% rule (a 95\% interval would contain both fits), so we read it as a pre-specified model-selection check, not a strong out-of-sample validation. At 15 Hz both fits predicted higher overshoot (62\% and 57\%) than the observed 7/15 = 47\%; that raw rate sits at the protocol-clean fit's lower 80\% bound and inside the sensitivity fit's interval, and is depressed by the non-arrival dilution described below --- we report it as pre-specified rather than setting it aside. Taken together, the pre-specified exercise favours the protocol-clean fit at 40 Hz while flagging that the slowest cell falls below both predictions for a documented reason.

\textbf{Non-arrival (premature) trials.} Distinct from overshoot, a fraction of continuous trials triggered the arrival rule \emph{before} the full footprint reached the top platform --- typically when a skewed approach placed one front foot on the platform while the others lagged, collapsing the body-pitch transient early. Because such a trial never reached the platform edge, it presents no opportunity to overshoot; we record it as a separate \emph{non-arrival} outcome rather than as a safe arrival or an overshoot. Non-arrivals occur in both arms at comparable rates (for example 4/15 continuous and 3/15 climb--settle at $\approx$15 Hz) and therefore do not bias the cadence contrast, but they are concentrated enough at $\approx$15 Hz to dilute that cell's raw continuous overshoot rate below the $\approx$20 Hz cell; restricting attention to trials that actually reached the summit decision restores a monotonic ordering across rates. We report the raw rates as the primary analysis and this conditional view as a supporting observation, and we do not remove non-arrival trials from the denominator of the primary analysis. The same trials were independently flagged by the blind video coder of Section 7.4 --- who, without knowledge of condition, scored them as neither clean arrivals nor overshoots --- an external check on this outcome category.

\begin{figure}[t]\centering
\includegraphics[width=\linewidth]{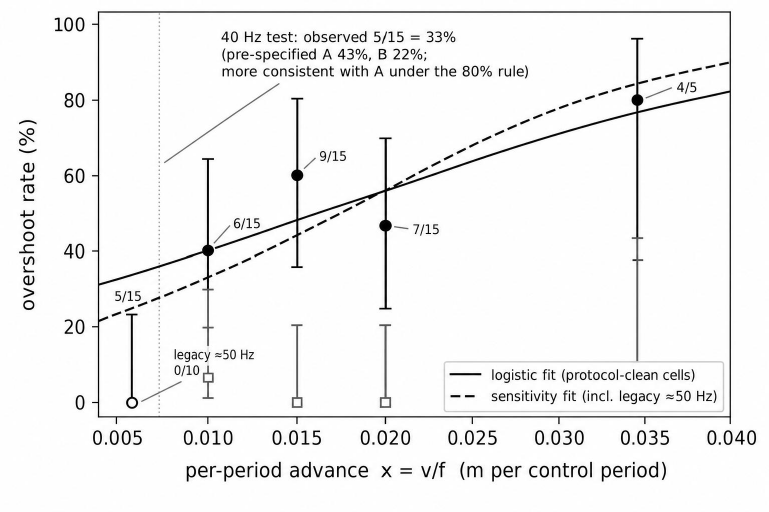}
\caption{\textbf{Latency dose--response of in-motion summit detection.} Per-cell overshoot rates of the continuous arm (markers with Wilson 95\% CIs) against the per-period advance x = v/f; curves: logistic fit on the protocol-clean cells (solid) and the sensitivity fit including the legacy $\approx$50 Hz diagnostic batch (dashed, open marker). Squares: climb--settle arm at the same doses (0--7\% throughout). The pre-specified 40 Hz validation dose is marked; the observed cell (5/15 = 33\%) is consistent with the protocol-clean fit under the pre-specified 80\% rule (Section 6.4).}
\end{figure}

Figure 9 shows the dynamics behind the rate. In a representative continuous trial at the deployed $\approx$15 Hz rate the body pitches deep during the climb and then swings nose-down past +12$^\circ$ as the robot crosses the top edge, while its forward displacement keeps growing --- it walks off the platform. In the matched climb--settle trial the body recovers toward level and holds, and the forward displacement plateaus at the summit; its stepped displacement profile is the cadence itself, each pause holding the robot while the rate-limited loop registers the arrival condition before further forward motion is commanded.
\begin{figure}[t]\centering
\includegraphics[width=\linewidth]{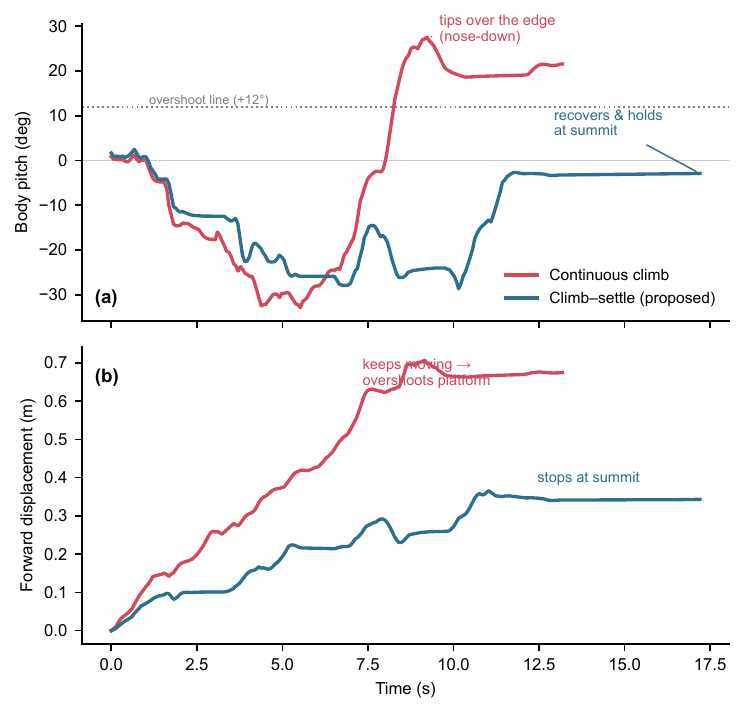}
\caption{\textbf{Representative overshoot dynamics at the deployed loop rate} ($\sim$15 Hz). (a) Body pitch: the continuous trial swings nose-down past +12$^\circ$ as the robot tips over the top edge, whereas the climb--settle trial recovers toward level and holds. (b) Forward displacement: the continuous trial keeps advancing and overshoots, whereas the climb--settle trial stops at the summit.}
\end{figure}

\textbf{Foot-force dwell.} The four foot-force channels give a geometric read on whether the full footprint has settled on the platform --- a question that matters because the top platform (50 $\times$ 50 cm) is shorter than the robot's standing length (Section 4.6). In the climb--settle trials a long, stable four-foot dwell is recorded ($\approx$4 s for the representative $\approx$15 Hz trial; Figure 10), whereas the continuous-overshoot trials show no sustained four-foot dwell (below 0.1 s; Figure 10). The dwell is one component of the overshoot criterion (Section 5.4), so it is not independent evidence of the overshoot outcome; it is computed from a channel the pitch-based controller does not use, and Figure 10 displays it to show directly that the cadence lands a stable four-foot stance while continuous overshoot does not. The long dwell is also partly a mechanical consequence of the climb--settle hold, a further reason to read Figure 10 as descriptive. The four-foot dwell is computed per trial across the whole set, not only for the representative pair of Figure 10; it is threshold-sensitive --- in several stably parked trials one foot rests just below the 24-unit contact threshold, so the all-four dwell understates a visibly stable stance --- which is a further reason we treat it as descriptive corroboration rather than the basis of the overshoot label, the rates in Table 8 resting on the full-sample pitch criterion.
\begin{figure}[t]\centering
\includegraphics[width=\linewidth]{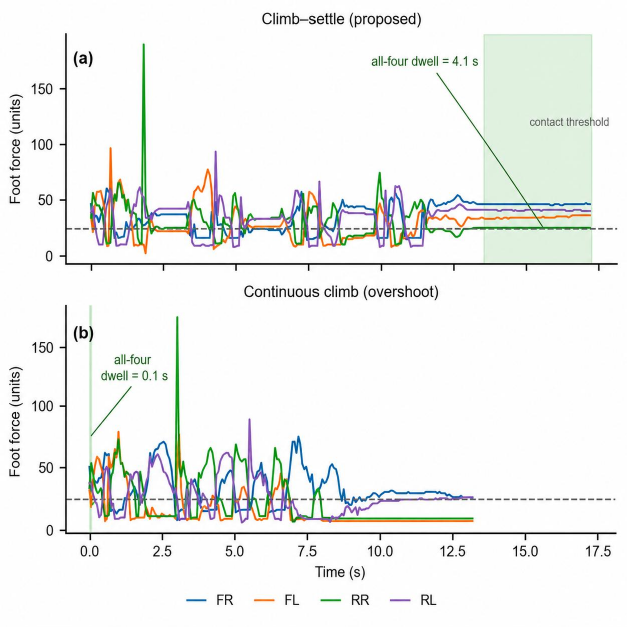}
\caption{\textbf{Four-foot-force dwell on the top platform.} (a) Climb--settle: all four channels stay above the contact threshold (24 units) for a long, stable dwell ($\sim$4 s). (b) Continuous (overshoot): no sustained four-foot dwell. The dwell is one component of the overshoot criterion (Section 5.4), computed from a channel the pitch-based controller does not use.}
\end{figure}

\textbf{Scope of the ablation.} This ablation isolates the climb--settle cadence --- the factor that addresses control-loop latency --- while holding the relative pitch-hysteresis arrival rule fixed in both arms. It therefore establishes that the cadence removes the overshoot failure of in-motion detection at the deployed loop rate; it does \textbf{not} separately ablate the arrival rule against a single absolute threshold, nor does it optimise the cadence parameters (1.0 s climb / 2.0 s stop). Those comparisons are left to future work (Section 7.4). One further scope condition is observed in the data: on this stepped platform the climb produces a \emph{single} continuous nose-up pitch excursion rather than separable per-level pitch peaks, so the arrival decision rests on the depth-and-recovery of that single excursion; counting per-level pitch events is discussed in Section 7 as a conditional mechanism for staircases whose geometry produces separable per-level transients.

\subsection{Integrated State-Machine Robustness}

The end-to-end and module-level campaigns together show that the state machine preserves the intended ordering of behaviours: route following, obstacle / corridor traversal, stair traversal, and post-stair continuation. No end-to-end trial failure was logged as a corridor wall contact or a stair-detection failure; both end-to-end failures occurred after the core corridor and stair behaviours had already been exercised, pointing to late-stage line / platform alignment as the next engineering bottleneck.

Overall the empirical pattern is coherent across scales. The isolated line follower and the split-turn corridor maneuver are stable; the stair ablation identifies a clear failure mode --- overshoot of the small top platform under control-loop latency in continuous detection --- and shows that the climb--settle cadence removes it at the deployed loop rate; and the full integrated route reaches 90.0\% completion within a mean runtime below 80 s. These results support the paper's bounded claim: a structured inspection route can be executed by a deterministic, onboard, low-sensing navigation stack when the route geometry is known and the behaviour-switching logic is engineered around the specific failure modes of each segment.

\section{Discussion}

The stair result motivates a mechanism-level interpretation that may be relevant beyond this particular system. On the evaluated compute-constrained platform, the navigation loop is slow relative to forward motion, so a detector that decides while the robot keeps moving can register a summit-arrival condition too late and overshoot a top platform shorter than the robot. Changing the cadence --- pausing so that the rate-limited loop can evaluate the arrival condition before further forward motion --- suppressed that failure in the tested conditions. The subsections below develop this bounded interpretation and its limits.

\subsection{What mapping-free and learning-free navigation can and cannot mean}

The descriptor \emph{mapping-free and learning-free navigation stack} used throughout this paper is precise about what is excluded and consequently about the scope of the claim. Three observations bound the claim. First, the exclusion set is concrete: 3D LiDAR point clouds, RGB-D and stereo depth, semantic segmentation networks, end-to-end learned-policy controllers, and any off-board computation are all absent from the stack. The manufacturer's closed-source base locomotion controller is treated as platform firmware: \emph{learning-free} refers to the authors' navigation and state-machine layer, which contains no learned policy, and makes no claim about the vendor's low-level gait controller. Second, the sensing modalities that \emph{are} used --- the built-in IMU, four foot-force sensors, three 1D range topics, and a monocular camera consumed by the line-extraction module --- are all onboard and either proprioceptive or single-task perceptive. The monocular camera is not an exception to the descriptor; it is a focused single-purpose sensor of the kind that the descriptor (mapping-free, learning-free) explicitly admits. Third, the system runs onboard on a Jetson Orin without SLAM, without a global map, and without policy-inference frameworks. The claim should therefore be read as: \emph{under the stated exclusion set and in the structured-inspection context, the navigation layer executed the calibrated route end-to-end in the reported trials.} It should not be read as a claim about navigation under arbitrary sensor exclusions, in arbitrary environments, or on routes for which the system has not been parameterized.

\subsection{Why the system is useful despite not being a general planner}

The integrated stack does not attempt to be a general navigation planner. The corridor maneuver is a deployable primitive for the specific geometric regime in which the corridor width approaches the body footprint diagonal; the stair detector assumes a multi-level stepped platform of approximately known geometry; the upper-level state machine assumes a fixed task chain (line $\rightarrow$ obstacle/corridor $\rightarrow$ stairs). This restriction is compatible with structured inspection, where repeated routes inside an HVDC converter station \cite{gehring2021field}, an oil-and-gas terminal \cite{bellicoso2018advances,anybotics_oilgas}, or a power plant \cite{bd_nationalgrid,bd_jpower,bd_austria} may be geometrically stable and known in advance. The cost, payload, and compute barriers identified by inspection survey work \cite{halder2023construction,ha2023robots} are also concentrated in the perceptive sensor and processing layer, which the present system deliberately reduces. The specific geometries evaluated here --- the 55 cm corridor and the 50 cm top platform --- are structured-task abstractions derived from a competition course (Section 5.1), not replicas of a measured inspection site; they instantiate classes of constraints rather than a documented field geometry.

\subsection{The mechanism-class argument for the stair detector}

Section 6.4 reports the central ablation of the paper, and Section 4.4 motivates its interpretation. The proposed stair detector is not offered as an \emph{optimally tuned} arrival threshold; its robustness in these trials comes from the \emph{climb--settle cadence} that precedes the arrival rule. On the evaluated platform, the navigation loop shares the processor with perception pipelines and runs at $\approx$15 Hz in the integrated stack. A continuous-climb detector can therefore register the body-pitch recovery after the robot has crossed the far edge of the short top platform (Section 6.4, Figure 9). The climb--settle cadence suppresses this observed failure by holding the robot stationary while the rate-limited loop evaluates the deep-climb-then-recovery condition. The same-speed $\approx$40$\rightarrow$$\approx$30$\rightarrow$$\approx$20 Hz steps support the proposed rate mechanism (33$\rightarrow$40$\rightarrow$60\% overshoot), and the cadence arm remains near zero across the tested rates (Table 8, Figure 7). The inference remains bounded: samples are moderate ($n=15$ per matrix cell), the evidence covers one stepped-platform geometry, robot, and operator, and the experiment isolates the cadence without separately ablating the arrival rule. Generalization to other platforms and loop architectures is therefore a testable hypothesis, not an established result.

Two further points bound the argument. First, the foot-force dwell (Figure 10) is part of the overshoot criterion (Section 5.4), not independent evidence of it; it is computed from a channel the controller does not use, and it shows directly that the cadence settles a stable four-foot stance whereas continuous overshoot does not. We deliberately do not treat it as independent confirmation: besides entering the overshoot label, the long dwell is partly a mechanical consequence of the climb--settle hold --- the robot stops and holds on the platform by design, whereas a trial that overshoots never holds. Second, the mechanism relies on the depth-and-recovery of a \emph{single} continuous pitch excursion (the recorded peak-event count at trigger is 1), which is why it does not depend on resolving separate per-level pitch peaks; on stepped platforms tall enough to produce separable per-level transients, an event-counting variant --- and, for in-loop enforcement of a small-platform constraint, the all-four-feet confirmation gate of the earlier exploratory variant --- become natural extensions (Section 7.5).

Three simpler alternatives to the cadence deserve comment, since it is one fix among several rather than the only one available. First, \emph{climbing more slowly} in continuous mode would also shorten the forward distance covered per loop iteration and thus reduce overshoot; the cadence is preferred not because slowing is ineffective but because it decouples the arrival decision from the loop rate while letting the robot retain a useful climb speed --- a slow continuous crawl trades throughput for safety, whereas the cadence does not. We did not run a slow-continuous arm and therefore cannot quantify that trade; it is a natural additional comparison. Second, \emph{exteroceptive edge detection} is not available in this configuration: the three 1D range beams face forward and laterally to detect stair entry (Section 4.4) and do not look toward the far platform edge, so they cannot sense the overshoot boundary. Third, a \emph{foot-force confirmation gate} on the summit transition was implemented in an earlier variant and dropped (Section 4.6); the dwell recomputation here supports that choice --- the four-foot contact signal is noisy and threshold-sensitive (in several stably parked trials one foot rests below the 24-unit contact threshold), so gating the control decision on it would be less reliable than the pitch-hysteresis rule. The cadence is therefore a deliberate selection among these alternatives, not the only mechanism tried.

\subsection{Limitations}

The system has several limitations that should be reported transparently. First, the physical course is structurally fixed: corridor width, stair geometry (riser/tread dimensions and the per-level platform extents of the three-level stepped platform), and line-track parameters are all known in advance and the upper-level state machine is hand-parameterized to them. Non-rectangular corridors, corridors with internal obstacles, or stepped platforms whose level geometry differs from the configured platform-size values are not handled by the current primitives without re-parameterization. Second, no SLAM or global map is constructed; the system does not know its absolute pose at any point during the run and cannot, for example, return to its starting position after the course terminates. Third, no outdoor or unstructured-terrain validation is reported; the system is targeted at indoor structured inspection and we make no claim about its behavior on natural terrain. Fourth, no long-duration endurance evaluation is reported; the longest run reported is the integrated course (85.57 s), and we do not characterize battery, sensor drift, or thermal effects beyond that horizon. Fifth, as discussed in Section 7.1, the line-following segment requires a painted track and the front-mounted monocular camera; the mapping-free and learning-free claim does not extend to navigation in environments that lack any visual cue and where no painted track is available.

A further group of limitations concerns the stair-detector evaluation specifically, and we state them plainly because they bound how the central ablation should be read. (i) \emph{The ablation isolates the cadence, not the arrival rule.} The proposed and continuous arms share the relative pitch-hysteresis arrival rule and differ only in the climb--settle cadence; the experiment therefore attributes the overshoot result to the cadence, but does not separately ablate the arrival rule against a single absolute threshold, nor does it optimise the cadence parameters (1.0 s climb / 2.0 s stop). (ii) \emph{Sample sizes are moderate.} Each protocol cell contains n = 15 trials; the per-cell Wilson intervals remain wide, but at $\approx$20 and $\approx$15 Hz the continuous and climb--settle cells now have separated intervals, and the aggregate cadence effect is established by the pooled and stratified tests of Section 6.4 rather than by any single cell. The design is fully crossed at $\approx$30/20/15 Hz, with an additional 40 Hz continuous validation cell. Two continuous trials whose per-iteration traces failed to record are omitted from the trace-based labelling, the same objective rule applied to both arms. (iii) \emph{Overshoot labels are trace-sourced and blind-validated.} The overshoot classification is computed from the logged pitch and foot-force traces by a fixed objective criterion (Section 5.4). An independent coder, blind to the trace data and to the experimental condition, labeled all 76 newly recorded trials from video; on the 61 trials the coder scored as safe or overshoot, agreement with the trace criterion is Cohen's $\kappa$ = 0.82 (observed agreement 92\%), an almost-perfect level. All five disagreements run the same direction --- the blind coder scored an overshoot where the more conservative trace criterion scored safe, all in the continuous arm --- so the criterion under-counts continuous overshoots rather than inflating them; one of the five is the operator-assisted $\approx$15 Hz trial whose manual hold the dwell conjunct had masked, which the blind coder independently caught. We retain the fixed-criterion labels without re-judging and report the disagreements. We additionally audited the climb--settle arm for the same contamination: no climb--settle trial involved an operator catch, and the only climb--settle trials with a pitch excursion past 12$^\circ$ either register as overshoots under the criterion or were independently scored safe by the blind coder, so no climb--settle overshoot is masked by a manual hold. The pitch-only robustness check --- dropping the dwell conjunct of the criterion across the full trace set --- has been carried out (Section 6.4): it can only hold or raise the climb--settle counts (it raises the $\approx$30 Hz cadence cell from 1/15 to 2/15, leaving the other cadence cells unchanged) and recovers the masked $\approx$15 Hz overshoot (continuous $\approx$15 Hz, 7/15 $\rightarrow$ 8/15); the cadence effect survives it. (iv) \emph{Single-factor generalization is untested:} all stair trials used one stepped-pyramid geometry, one robot, and one operator, so robustness to geometry, platform, and operator variation is not characterized. (v) \emph{The legacy $\approx$10 Hz cells carry a different climb speed.} They were run at 0.35 m/s rather than the uniform 0.30 m/s of the main matrix, and are reported only as a higher-dose sensitivity condition; the speed difference is carried explicitly by the dose variable x = v/f of the model (Section 4.5) rather than entering the matched 0.30 m/s $\approx$30/20/15 Hz matrix.

\subsection{Deployment and reproducibility implications}

The two design choices that most directly enable practical deployment are the onboard runtime (no off-board compute or network dependency) and the minimal sensor budget (no LiDAR point cloud, RGB-D, or learned-policy inference engine). Both choices reduce integration demands on a fielded platform. The C++ / \texttt{unitree\_sdk2} reference stack is available from the corresponding author on reasonable request and is intended for public release after repository cleanup; the parameter values reported in Table 2 support replication on a Unitree Go2 EDU with a structurally similar physical track. Future work includes: (i) automated geometry parameterization, (ii) extension of the upper-level state machine beyond the fixed line $\rightarrow$ obstacle $\rightarrow$ stair task chain, (iii) richer inspection-task integration \cite{gehring2021field}, (iv) a longer-duration endurance study on a realistic inspection loop, and (v) evaluating a four-foot confirmation gate for deployments in which the small-platform constraint must be enforced during control rather than verified afterwards.

\section{Conclusions}

This paper began with a deployment failure: on a compute-constrained quadruped whose navigation loop shares the processor with perception pipelines, declaring stair-summit arrival from body pitch while the robot keeps climbing can register the condition too late for a top platform shorter than the robot. A controlled cadence $\times$ loop-rate ablation ($n=15$ per matrix cell) showed that continuous-climb overshoot increased with per-period advance $x=v/f$, whereas the climb--settle cadence held the observed rate near zero across the tested conditions (pooled 22/45 vs 1/45 over $\approx$30/20/15 Hz; rate-stratified CMH $p\approx1.9 \times 10^{-6}$; pooled Fisher $p\approx2.4 \times 10^{-7}$). A logistic dose--response model summarizes the failure and gives a model-based critical rate of $\approx$19 Hz at 0.30 m/s for the tested geometry. The ablation isolates the cadence; the arrival rule was held fixed in both arms, and non-arrival outcomes are reported separately from overshoot (Section 6.4).

The second result is geometric: contact-free in-place 90$^\circ$ yaw of the nominal swept envelope in a corridor narrower than $\approx0.85\sqrt{L^{2}+W^{2}}$ is infeasible even when the corner junction is used, and the three-segment maneuver reduces the computed width requirement by $\approx$20\%. Hardware trials corroborate the analysis (20/20 contact-free versus 14/20 completions with 12 wall contacts; exit-heading error 1.56$\pm$0.71$^\circ$ versus 5.64$\pm$2.50$^\circ$, Mann--Whitney $p\approx1.0 \times 10^{-5}$). Both primitives were integrated in a fully onboard, mapping-free and learning-free navigation stack using the built-in IMU, four foot-force channels, three 1-D ranges, and one line camera; the stack completed the calibrated physical inspection course in 18 of 20 trials (90.0\%).

The system is bounded but useful: it is targeted at the structured-inspection deployment context and we make no claim of generality beyond that envelope. Future work includes automated geometry parameterization, extension of the upper-level state machine beyond the fixed task chain reported here, richer inspection-task integration \cite{gehring2021field}, and longer-duration endurance evaluation on a realistic inspection loop.

\backmatter

\section*{Statements and Declarations}

\begin{itemize}
\item \textbf{Funding.} The authors declare that no funds, grants, or other support were received during the preparation of this manuscript.
\item \textbf{Competing interests.} The authors declare no competing interests.
\item \textbf{Ethics approval and consent to participate.} Not applicable.
\item \textbf{Consent for publication.} Not applicable.
\item \textbf{Data availability.} The per-trial raw onboard traces (body pitch, four-channel foot force, 1-D range, and timing) and representative trial recordings supporting the findings are available from the corresponding author on reasonable request. The authors intend to deposit a citable public archive upon acceptance.
\item \textbf{Code availability.} The analysis and figure-generation scripts are available from the corresponding author on reasonable request. The integrated C++ / \texttt{unitree\_sdk2} reference stack will be released in a public repository after repository cleanup.
\item \textbf{Author contributions.} Hanting Suo and Haonan Yan contributed equally to this work: they designed and implemented the navigation stack and the analysis pipeline, performed the experiments, curated the data, and wrote the original draft. Liang Wang contributed to the methodology and provided supervision. Aiguo Song conceived and supervised the project, provided resources, and reviewed and edited the manuscript. All authors read and approved the final manuscript.
\end{itemize}

\begingroup
\emergencystretch=3em
\bibliography{refs}
\endgroup

\end{document}